\documentclass[runningheads]{llncs}
\usepackage{graphicx}
\usepackage{amsmath,amssymb} 
\usepackage{color}
\usepackage[width=122mm,left=12mm,paperwidth=146mm,height=193mm,top=12mm,paperheight=217mm]{geometry}
\usepackage{cite}
\newcommand\blfootnote[1]{%
  \begingroup
  \renewcommand\thefootnote{}\footnote{#1}%
  \addtocounter{footnote}{-1}%
  \endgroup
}
\begin{document}
\mainmatter
\def\ECCV16SubNumber{***}  

\title{Computer Vision with Deep Learning for Plant Phenotyping in Agriculture: A Survey} 



\author{\hspace{-5mm}Akshay L Chandra$^{\dagger 1}$, Sai Vikas Desai$^{\dagger 1}$, \hspace{15mm} Wei Guo$^2$ \newline \hspace{-40mm} Vineeth N Balasubramanian$^1$  }


\institute{\textbf{\hspace{10mm}$^1$Indian Institute of Technology \hspace{10mm} $^2$The University of Tokyo \newline \hspace{-45mm} Hyderabad}}

\maketitle

\begin{abstract}
In light of growing challenges in agriculture with ever growing food demand across the world, efficient crop management techniques are necessary to increase crop yield. Precision agriculture techniques allow the stakeholders to make effective and customized crop management decisions based on data gathered from monitoring crop environments. Plant phenotyping techniques play a major role in accurate crop monitoring. Advancements in deep learning have made previously difficult phenotyping tasks possible. This survey aims to introduce the reader to the state of the art research in deep plant phenotyping.

\end{abstract}

\blfootnote{
$\dagger$Equal Contribution}

\section{Introduction}

Population growth, increasing incomes, and rapid urbanization in developing countries are expected to cause a drastic hike \cite{Agro2050} in food demand. This projected rise in food demand poses several challenges to agriculture. Owing to a continuous decline in global cultivable land \cite{arable}, increasing the productivity of the existing agricultural land is highly necessary. This need has led to the scientific community focusing their efforts \cite{sust1, sust2, sust3} on developing efficient and sustainable ways to increase crop yield. To this end, precision agriculture techniques have attracted a lot of attention. Precision agriculture is a set of methods to monitor crops, gather data, and carry out informed crop management tasks such as applying the optimum amount of water, selecting suitable pesticides, and reducing environmental impact. These methods involve the usage of specialized devices such as sensors, UAVs, and static cameras to monitor the crops. Accurate crop monitoring goes a long way in assisting farmers in making the right choices to obtain the maximum yield. Plant phenotyping, a rapidly emerging research area, plays a significant role in understanding crop-related traits. Plant phenotyping is the science of characterizing and quantifying the physical and physiological traits of a plant. It provides a quantitative assessment of the plant's properties and its behavior in various environmental conditions. Understanding these properties is crucial in performing effective crop management. 

Research in plant phenotyping has grown rapidly thanks to the availability of cost-effective and easy to use digital imaging devices such as RGB, multispectral, and hyperspectral cameras, which have facilitated the collection of large amounts of data. This influx of data coupled with the usage of machine learning algorithms has fueled the development of various high throughput phenotyping tools [refs] for tasks such as weed detection, fruit/organ counting, disease detection and yield estimation. A machine learning pipeline typically consists of feature extraction followed by a classification/regression module for prediction. While machine learning techniques have helped build sophisticated phenotyping tools, they are known to lack robustness. They rely heavily on handcrafted feature extraction techniques and manual hyperparameter tuning methods. As a result, if feature extraction is not carefully done under a domain expert's supervision, they tend to perform poorly in uncontrolled environments such as agricultural fields where factors such as lighting, weather, exposure, etc. often cannot be regulated. Hence, feature extraction from data has been one of the major bottlenecks in the development of efficient high throughput plant phenotyping systems.

\begin{figure}
\centerline{\includegraphics[scale=0.7]{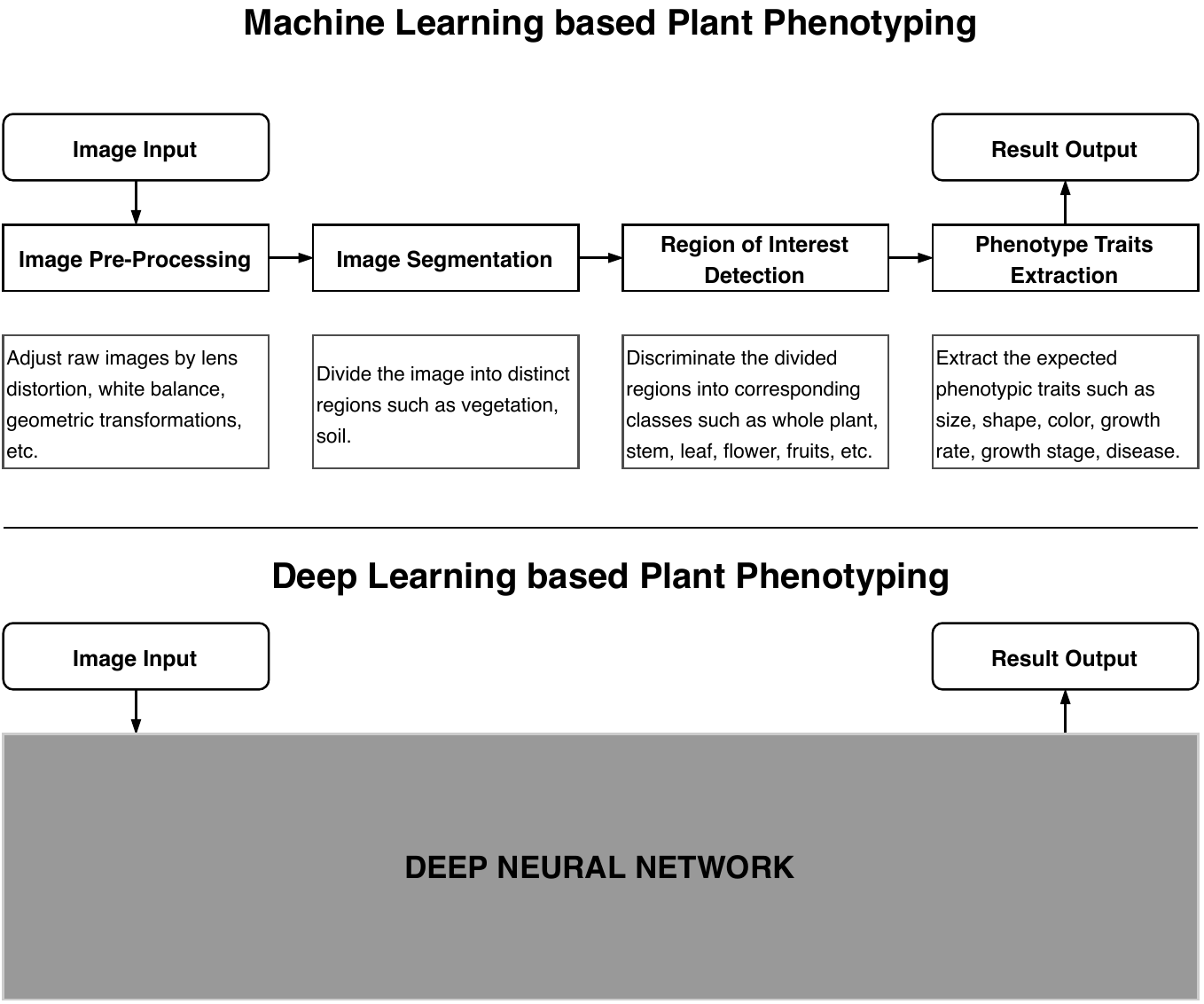}}
\caption{Difference between ML and DL based Plant Phenotyping.}
\label{fig-ml-dl}
\end{figure}

Advancements in deep learning, a sub-field of machine learning which allows for automatic feature extraction and prediction on large scale data, has led to a surge in the development of visual plant phenotyping methods. Deep learning is particularly well-known for its effectiveness in handling vision-based tasks such as image classification, object detection, semantic segmentation, and scene understanding. Coincidentally, many of these tasks form the backbone for various plant phenotyping tasks such as disease detection, fruit detection, and yield estimation. Fig. \ref{fig-ml-dl} illustrates the difference between machine learning based plant phenotyping and deep learning based plant phenotyping. We believe that the expressive power and robustness of deep learning systems can be effectively leveraged by plant researchers to identify complex patterns from raw data and devise efficient precision agriculture methodologies. The purpose of this survey is to enable the readers to get a bird's eye view of the advancements in the field of deep learning based plant phenotyping, understand the existing issues, and become familiar with some of the open problems which warrant further research.

\section{Background}

\subsection{Plant Phenotyping}

Plant phenotyping is the science of quantifying the physical and physiological traits of a plant. Plant phenotyping mainly benefits two communities: farmers and plant breeders. By better understanding the traits of the crop, a farmer can optimize crop yield by making informed crop management decisions. Similarly, understanding the crop's behavior is crucial for plant breeders to select the best possible crop variety for a given location and environment. In the past, plant phenotyping was a manual endeavor. The process of manually observing a small set of crop samples and reporting observations periodically was slow, labor intensive and inefficient. The low throughput nature of these methods has impeded the progress in plant breeding research. However, the advent of modern data acquisition methods with various sensors, cameras and UAVs (Unmanned Aerial Vehicles) coupled with advances in machine learning techniques have resulted in the development of high-throughput plant phenotyping methods to be effectively used for precision agriculture.  

\begin{figure}
\centerline{\includegraphics[scale=0.3]{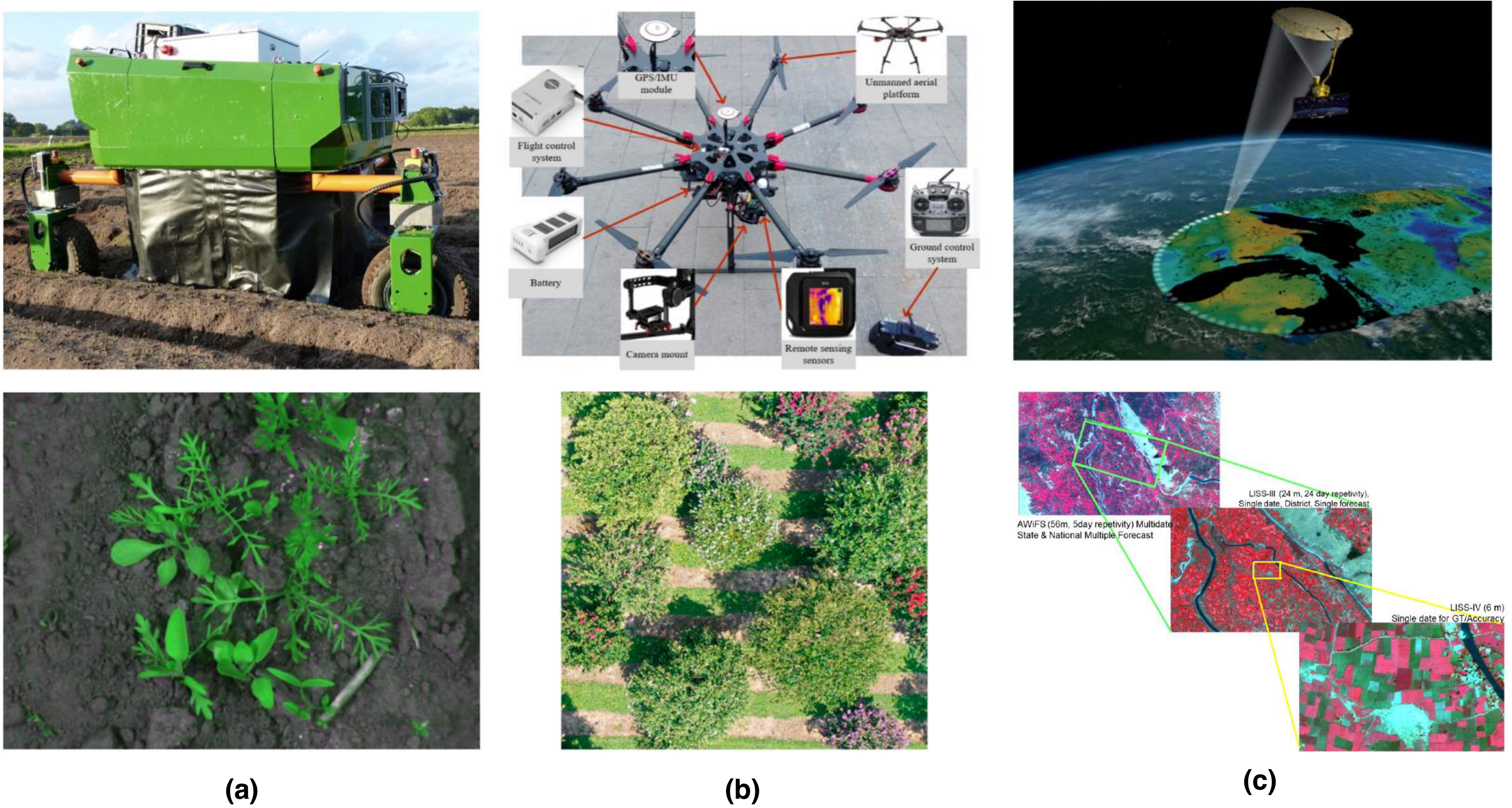}}
\caption{Top row of \textbf{(a)} shows BoniRob \cite{bonirob} a ground-based remote sensing robot, \textbf{(b)} shows an unmanned aerial vehicle \cite{miniUAV}, \textbf{(c)} shows a satellite scanning large areas of land respectively. Bottom row across \textbf{(a)}, \textbf{(b)}, and \textbf{(c)} shows corresponding example images acquired. Satellite Image Credits: NASA.}
\label{fig-three-modes}
\end{figure}

Depending on the method of data collection, plant phenotyping techniques can be classified into ground based, aerial and satellite based methods. In ground based phenotyping, high precision sensors are embedded in handheld devices or mounted on movable vehicles to measure useful traits such as plant height, plant biomass, crop development stage, crop yield etc. Fig. \ref{fig-three-modes} contracts the discussed classifcations. Movable phenotyping vehicles like BoniRob \cite{bonirob} have been developed where RGB cameras, hyperspectral cameras, LIDAR sensors, GPS receivers and other sensors can be mounted. Aerial based methods typically involve the usage of Unmanned Aerial Vehicles (UAVs) for crop monitoring. The recent advancements in UAVs and high resolution cameras have allowed the researchers to obtain high quality crop images. Tasks such as weed mapping, crop yield estimation, plant disease detection and pesticide spraying have been effectively carried out by UAVs. Satellite based plant phenotyping involves remote sensing of agricultural plots from satellites such as Landsat-8 and WorldView-3. Satellite based methods have been typically used for crop health monitoring over a large scale area such as a region/country. 
However, the cost of obtaining satellite images, the effect of clouds and the time gap between capturing and obtaining images inhibits its applicability for high throughput plant phenotyping in precision agriculture. 

\smallskip
With a variety of data collection tools at our disposal, large amounts of image and sensor data have been made available for plant phenotyping research. The next section introduces deep learning, a set of methods which can effectively recognize useful patterns in huge datasets. 
\subsection{Deep Learning}
\begin{figure}
\centerline{\includegraphics[scale=0.35]{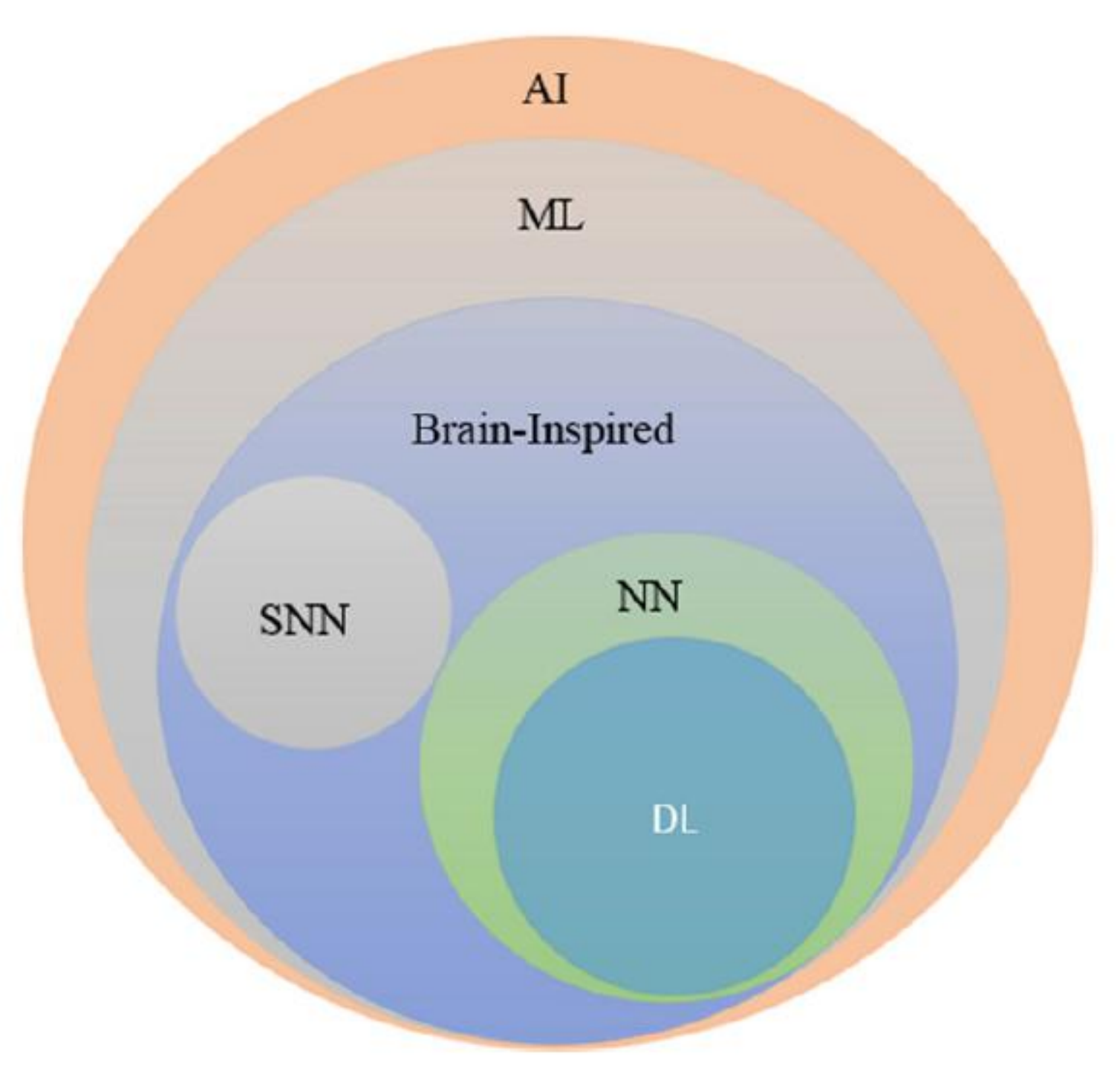}}
\caption{The taxonomy of AI \cite{Alom_2019}. AI: Artificial Intelligence; ML: Machine Learning; NN: Neural Networks; DL:Deep Learning; SNN: Spiking Neural Networks.}
\label{fig-taxonomy-of-ai}
\end{figure}

Machine Learning (ML) is a subset of Artificial Intelligence (AI), that deals with an algorithmic approach of learning from observational data without being explicitly programmed. ML has unimaginably revolutionized several fields in the last few decades. Neural Networks (NN) \cite{fitch_1944, Rosenblatt1958ThePA, universalapprox} is a sub-field of ML and it was this sub-field that spawned Deep Learning (DL). Among the most prominent factors that contributed to the huge boost of deep learning are the appearance of large, high-quality, publicly available labelled datasets, along with the empowerment of parallel GPU computing, which enabled the transition from CPU-based to GPU-based training thus allowing for signifcant acceleration in deep models’ training.  Since its redemption in 2006 \cite{unsuppretraining}, DL community has been creating ever more complex and intelligent algorithms, showing better than human performances in several intelligent tasks. The \textit{deep} in deep learning comes from the deep architectures of learning or the hierarchical nature of its algorithms. DL algorithms stack several layers of non-linear information processing units between input and output layer, called Artificial Neurons (AN). The stacking of these ANs in a hierarchical fashion allows for exploitation of feature learning and pattern recognition through efficient learning algorithms. It is proven that NNs are universal approximator of any function\cite{universalapprox}, making DL task agnostic \cite{taskagnosticbengio}. Fig. \ref{fig-taxonomy-of-ai} depicts the taxonomy of AI.   

Deep learning approaches may be categorized as follows: Supervised, semi-supervised or partially supervised, and unsupervised\footnote{Reinforcement Learning (RL) or Deep RL (DRL) is often treated as a semi-supervised or sometimes unsupervised approach.}. Supervised learning techniques use labeled data. In supervised DL, the environment includes sets of input and corresponding output pairs (often in large amounts), a criterion that evaluates model performance at all times called cost or loss function, an optimizing algorithm that minimizes the cost function with respect to the given data. Semi-supervised learning techniques use only partially labeled datasets (usually small amounts of label data, large amounts of unlabeled data). The popular Generative Adversarial Networks (GAN) \cite{GAN} are semi-supervised learning techniques. Unsupervised learning systems function without the presence of labeled data. In this case, the system learns the internal representation or important features to discover unknown relationships or structure within the input data. Often clustering, dimensionality reduction, and generative techniques are considered as unsupervised learning approaches.

\subsection{Deep Learning for Computer Vision}

\subsubsection*{Convolutional Neural Networks} 

\begin{figure}
\centerline{\includegraphics[scale=0.4]{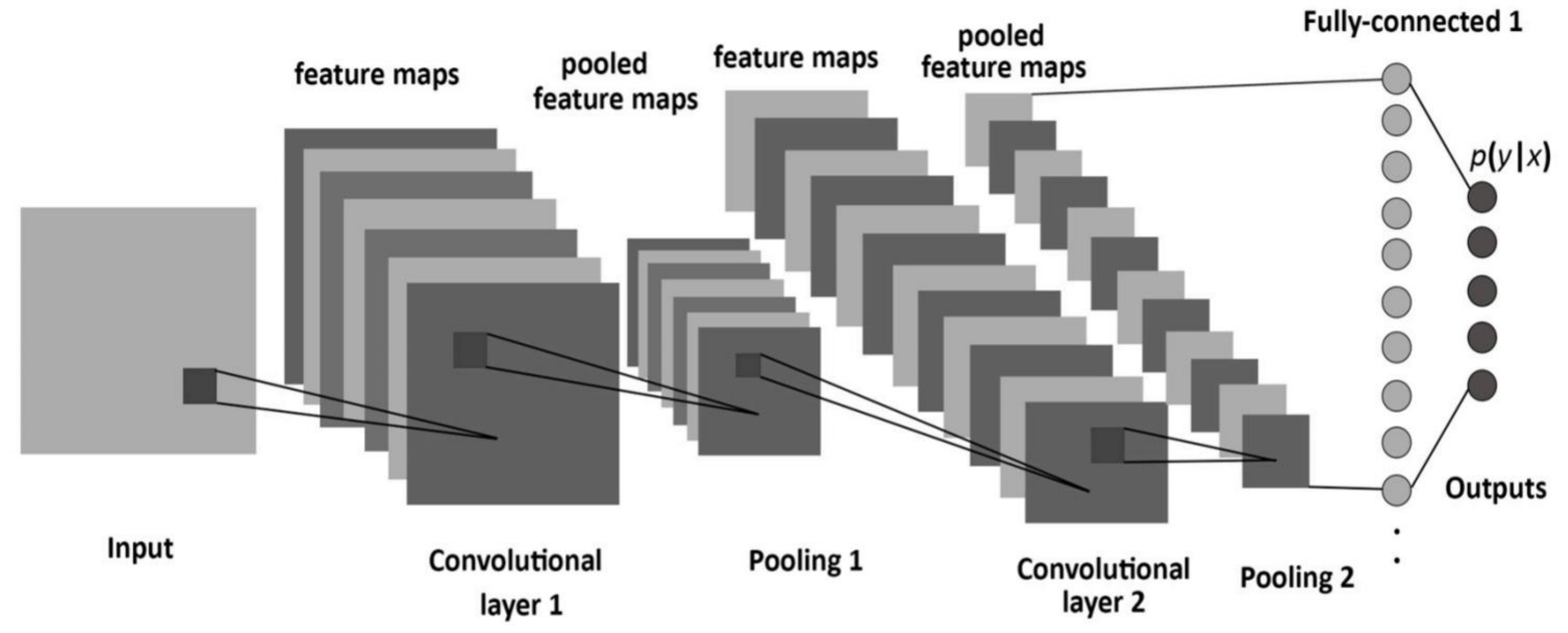}}
\caption{The structure of a CNN \cite{cnn-figure-ref}, consisting of convolutional, pooling, and fully-connected layers.}
\label{fig-cnn}
\end{figure}

Convolutional Neural Networks (CNN) is a subclass of neural networks that takes advantage of the spatial structure of the inputs. This network structure was first proposed by Fukushima in 1988 \cite{Fukushima1988NeocognitronAH}. It was not widely used then, however, due to limits of computation hardware for training the network. In the 1990s, LeCun et al. \cite{Lecun98Gradient} applied a gradient-based learning algorithm to CNNs and obtained successful results for the handwritten digit classification problem. CNNs have been extremely successful in computer vision applications, such as face recognition, object detection, powering vision in robotics, and self-driving cars. CNN models have a standard structure consisting of alternating convolutional layers and pooling layers (often each pooling layer is placed after a convolutional layer). The last layers are a small number of fully-connected layers, and the final layer is a softmax classifier as shown in Fig. \ref{fig-cnn}.  Every layer of a CNN transforms the input volume to an output volume of neuron activation, eventually leading to the final fully connected layers, resulting in a mapping of the input data to a 1D feature vector. In a nutshell, CNN comprises three main types of neural layers, namely, (i) convolutional layers, (ii) pooling layers, and (iii) fully connected layers. Each type of layer plays a diferent role.
\smallskip

\begin{figure}
\centerline{\includegraphics[scale=0.3]{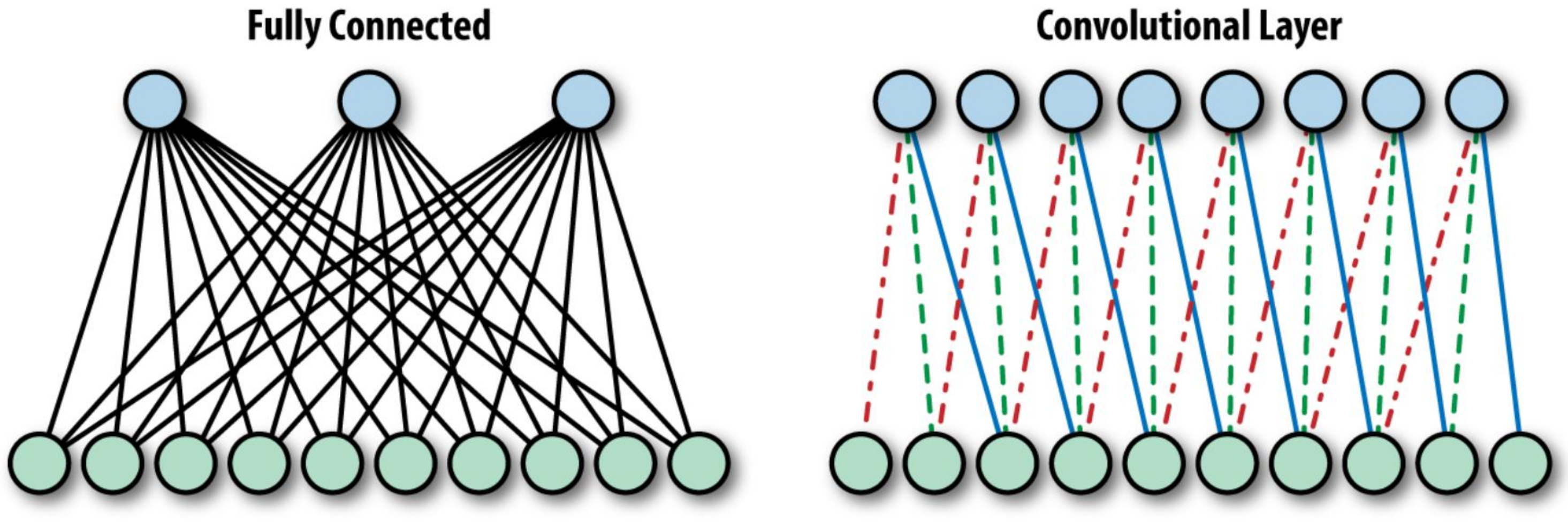}}
\caption{In a fully connected layer (left), each unit is connected to all units of the previous layers. In a convolutional layer (right), each unit is connected to a constant number of units in a local region of the previous layer. Furthermore, in a convolutional layer, the units all share the weights for these connections, as indicated by the shared linetypes. Figure and description are taken from \cite{fcnn-cnn-figure}.}
\label{fig-fcnn-cnn}
\end{figure}

\hspace{-5mm}\textit{(i) Convolution Layers}. In the convolutional layers, a CNN
convolves the whole image as well as the intermediate feature maps with different kernels, generating various feature maps. Exploiting the advantages of the convolution operation, several works have proposed it as a substitute for fully connected layers with a view to attaining faster learning times. Difference between a fully connected layer and a convolutional layer is shown in Fig. \ref{fig-fcnn-cnn}.        
\smallskip

\hspace{-5mm}\textit{(ii) Pooling Layers}. Pooling layers handle the reduction of the spatial dimensions of the input volume for the convolutional layers that immediately follow. The pooling layer does not affect the depth dimension of the volume. The operation performed by this layer is also called subsampling or downsampling, as the reduction of size leads to a simultaneous loss of information. However, such a loss is beneficial for the network because the network is forced to learn only meaningful feature representation. On top of that, the decrease in size leads to less computational overhead for the upcoming layers of the network, and also it works against overfitting. Average pooling and max pooling are the most commonly used strategies. In \cite{pooling-paper} a detailed theoretical analysis of max pooling and average pooling performances is given, whereas in \cite{pooling-paper2} it was shown that max pooling can lead to faster convergence, select superior invariant features, and improve generalization. 
\smallskip

\hspace{-5mm}\textit{(iii) Fully Connected Layers}.
Following several convolutional and pooling layers, the high-level reasoning in the neural network is performed via fully connected layers. Neurons in a fully connected layer have full connections to all activation in the previous layer, as their name implies. Their activation can hence be computed with a matrix multiplication followed by a bias offset. Fully connected layers eventually convert the 2D feature maps into a 1D feature vector. The learned vector representations either could be fed forward for classification or could be used as feature vectors for further processing.





\subsubsection*{Object Detection and Segmentation}

\begin{figure}
\centerline{\includegraphics[scale=0.2]{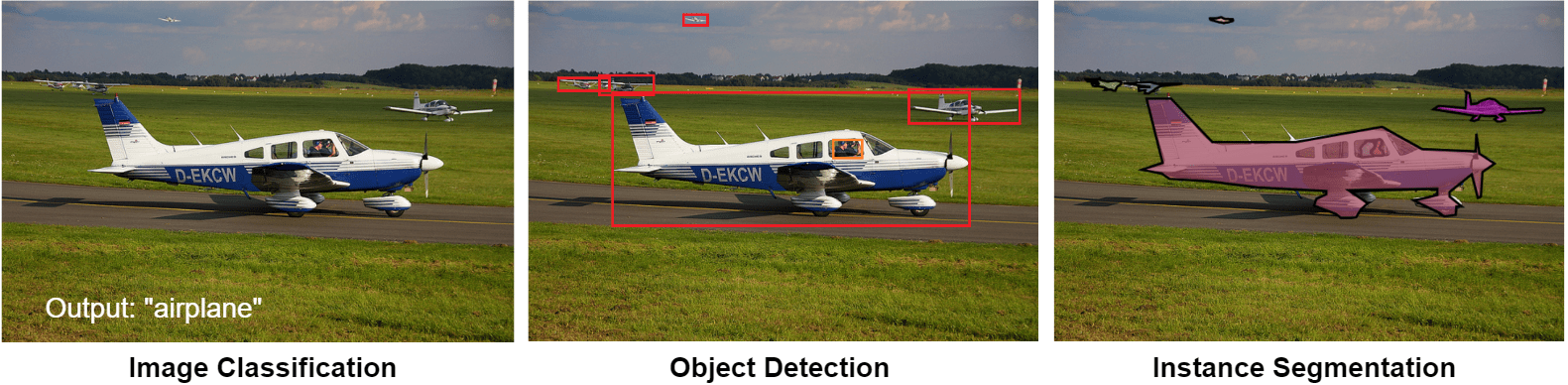}}
\caption{Visual illustration of difference between tasks - Image Classification, Object Detection and Instance Segmentation. Example taken from MS-COCO Dataset \cite{mscoco}.}
\label{fig-class-det-seg}
\end{figure}

Object detection and segmentation are two of the most important and challenging branches of computer vision, which have been widely applied in real-world applications, such as monitoring security, autonomous driving and so on, with the purpose of locating instances of semantic objects of a certain class. In a nutshell, object detection is the task of identifying locating objects (with bounding boxes) in images. While the task of segmentation is to classify each pixel of images with objects (dog, cat, airplane, etc.). We refer readers to \cite{det-survey, seg-survey} for more information on these tasks. Fig. \ref{fig-class-det-seg} visually contrasts the difference between these tasks.

\section{Application of Deep Learning in Plant Phenotyping}

\subsection{Ground-Based Remote Sensing for Plant Phenotyping}
Automation in agriculture and robotic precision agriculture activities demand a lot of information about the environment, the field, the condition and the phenotype of individual plants. An increase in availability of data allowed for successful usage of such robotic tools in real-world conditions. Taking advantage of the available data, combined with the availability of robots such as BoniRob \cite{bonirob} that navigate autonomously in fields, computer vision with deep learning has played a prominent role in realizing autonomous farming. Previously laborious jobs of actively tracking certain measurements of interest such as plant growth rate, plant stem position, biomass amount, leaf count, leaf area, inter crop spacing, crop plant count and others can now be done almost seamlessly.

\subsubsection*{Crop Identification and Classification}

A crucial prerequisite for selective and plant-specific treatments is that farming robots need to be equipped with an effective plant identification and classification system providing the robot with the information where and when to trigger its actuators to perform the desired action in real-time. For example, weeds generally have no useful value in terms of food, nutrition or medicine yet they have accelerated growth and parasitically compete with actual crops for nutrients and space. Inefficient processes such as hand weeding has
led to significant losses and increasing costs due to manual labour \cite{cropweed4}, which is why a lot of research is being done on crop vs weed classification and weed identification \cite{cropweed, cropweed2, cropweed3, cropweed5, weed-net, weed-id} and plant seedlings classification \cite{plant-seedlings1,plant-seedlings2}. This is extremely useful in improving the efficiency of precision farming techniques on weed control by modulating herbicide spraying appropriately to the level of weeds infestation.

\subsubsection*{Crop Detection and Segmentation}

\begin{figure}
\centerline{\includegraphics[scale=0.3]{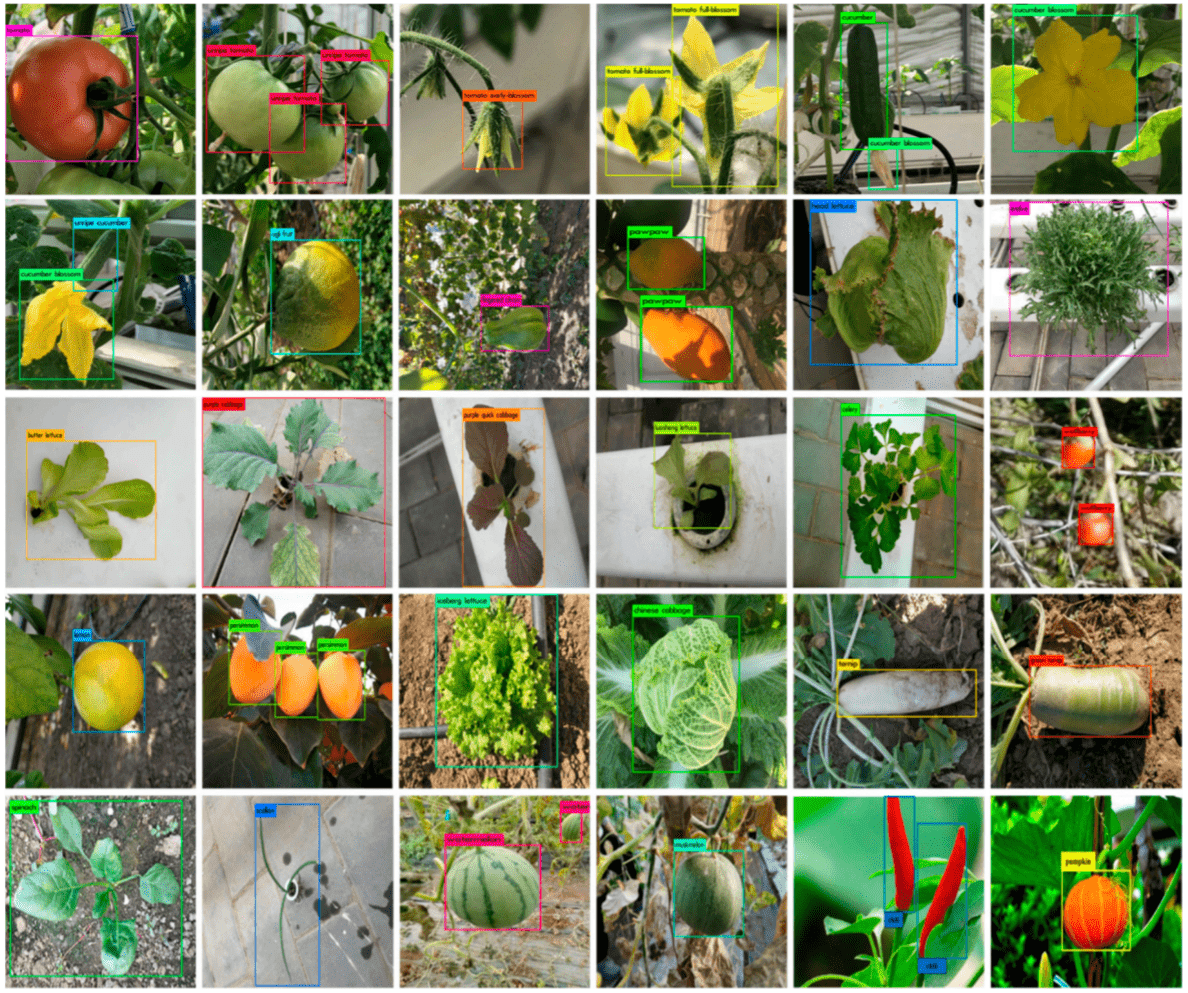}}
\caption{Some annotation examples from CropDeep dataset \cite{cropdeep}.}
\label{fig-cropdeep}
\end{figure}

\begin{figure}
\centerline{\includegraphics[scale=0.7]{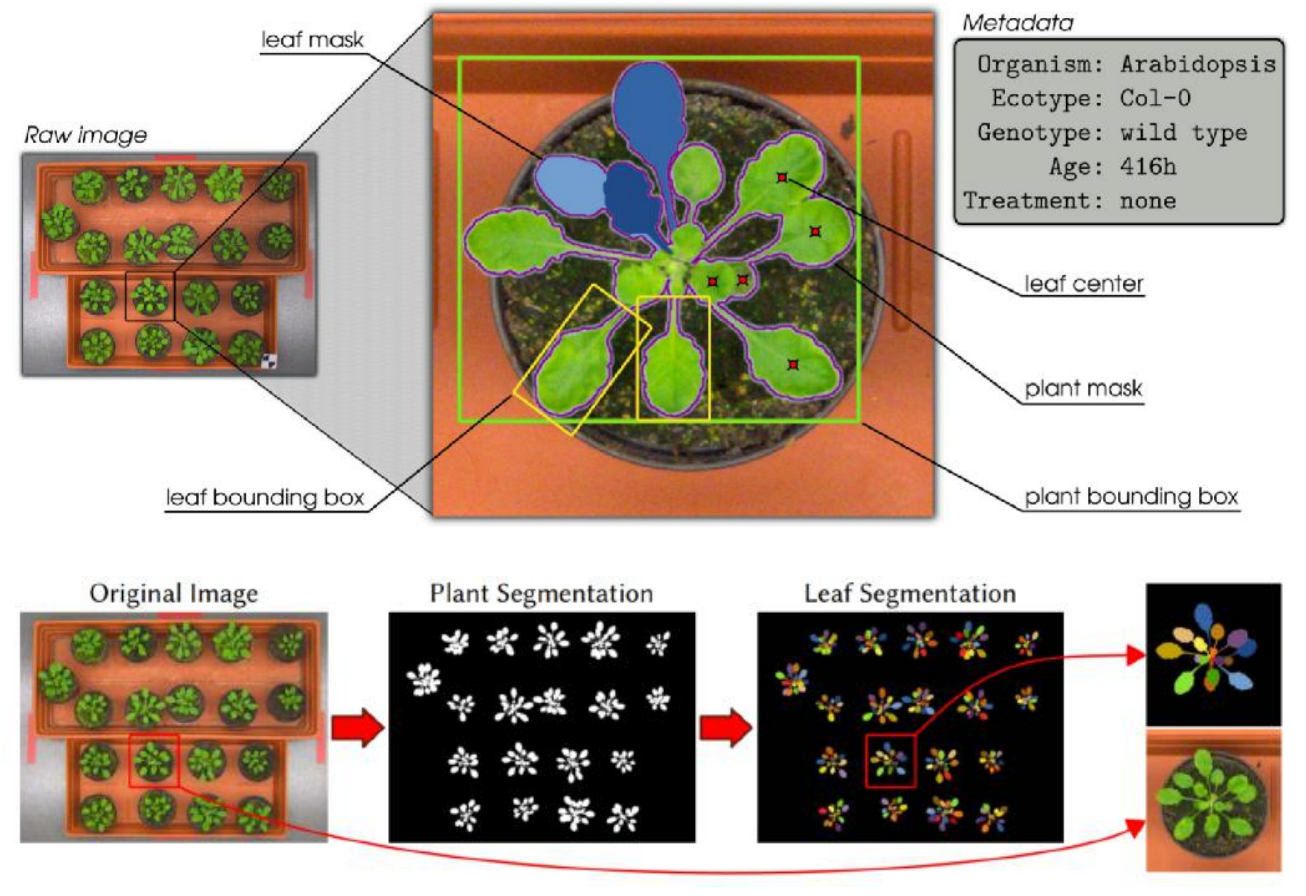}}
\caption{Visual illustration of all types of annotations available in \cite{rosette1, rosette2} dataset.}
\label{fig-rosette}
\end{figure}

Crop detection in the wild is arguably the most crucial step in the pipeline of several farm management tasks such as visual crop categorization \cite{crop-categorization}, real-time plant disease and pest recognition \cite{pest-recog-tomato}, picking and harvesting automatic robots \cite{harvesting-systems}, healthy and quality monitoring of crop growing \cite{irrigation-growth-monitor} and yield estimation \cite{sorghum-guo}. However, existing deep learning networks achieving state-of-the-art performance in other research fields are not suitable for agricultural tasks of crop management such as irrigation \cite{irrigation}, picking \cite{apple-picking}, pesticide spraying \cite{pesticide-spraying}, and fertilization \cite{fertilization}. The dominating cause is lack of diverse set of public benchmark datasets that are specifically designed for various agricultural missions. Some of the few rich datasets available are CropDeep \cite{cropdeep} for detection, multi-modal datasets like Rosette plant or \textit{Arabidopsis} datasets \cite{rosette1, rosette2, pp-dataset}, Sorghum-Head \cite{sorghum-guo}, Wheat-Panicle \cite{wheat-madec}, Crop/Weed segmentation \cite{cropweed}, and Crop/Tassle segmentation \cite{crop-tassel-dataset}. Fig. \ref{fig-cropdeep} contains some examples from the CropDeep \cite{cropdeep} dataset. Fig. \ref{fig-rosette} depicts multi-modal annotations provided in the Rosette Plant Phenotyping dataset \cite{rosette1, rosette2} i.e., annotations for detection, segmentation, leaf center along with otherwise rarely found meta data. 

Efficient yield estimation from images is also one of the key tasks for farmers and plant breeders to accurately quantify the overall throughput of their ecosystem. Recent efforts in panicle or spike detection \cite{weak-sup1, Desai2019, wspike, sorghum-guo}, leaf counting \cite{leaf_counting}, fruit detection \cite{Sa2016DeepFruitsAF} as well as pixel-wise segmentation-based tasks such as panicle segmentation \cite{panicle_seg,Oh2019} show very promising results in this direction. 

\subsubsection*{Crop Disease and Pest Recognition}

Modern technologies have given human society the ability to produce enough food to meet the demand of more than 7 billion people. However, food security remains threatened by a number of factors including climate change \cite{food-security}, the decline in pollinators \cite{pollinators}, plant diseases \cite{plant-disease}, and others. Plant diseases are not only a threat to food security at the global scale, but can also have disastrous consequences for smallholder farmers whose livelihoods depend on healthy crops. India loses 35\% of the annual crop yield due to plant diseases \cite{plantdoc}. In the developing world, more than 80 percent of the agricultural production is generated by smallholder farmers \cite{smallholders}, and reports of yield loss of more than 50\% due to pests and diseases are frequent \cite{madagascar}. Furthermore, the largest fraction of hungry people (50\%) live in smallholder farming households \cite{world-hunger}, making smallholder farmers a group that’s particularly vulnerable to pathogen-derived disruptions in food supply.

\begin{figure}
\centerline{\includegraphics[scale=0.3]{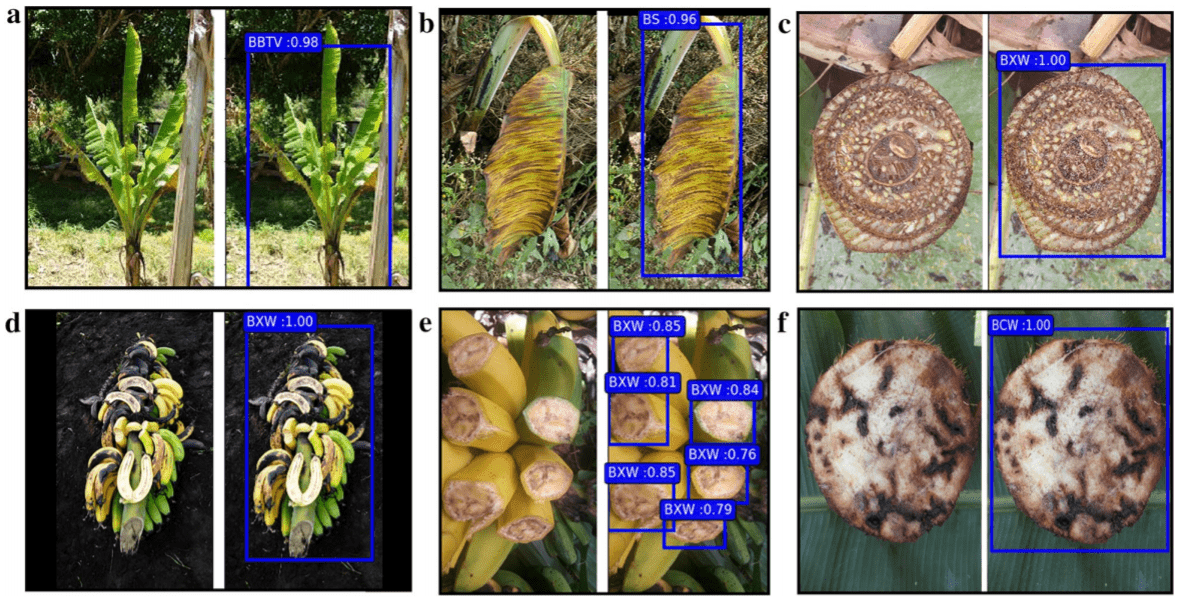}}
\caption{Detected classes and expected output of the trained disease detection model. \textbf{a} Entire plant afected by banana bunchy top virus (BBTV), \textbf{b} leaves affected by black sigatoka (BS), \textbf{c} cut pseudostem of Xanthomonas wilt (BXW) afected plant showing yellow bacterial ooze, \textbf{d} fruit bunch afected by Xanthomonas wilt (BXW), e cut fruit afected by Xanthomonas wilt (BXW), \textbf{f} corm afected by banana corm weevil (BCW). Figure and description taken from \cite{banana}.}
\label{fig-disease}
\end{figure}

Owing to these factors, timely disease and pest recognition becomes a priority task for farmers. In addition to that, farmers do not have many options other than consulting other fellow farmers or seeking help from government funded helplines \cite{kisan-helpline}. Availability of public datasets such as PlantVillage \cite{plant-village}, PlantDoc \cite{plantdoc} allowed for progress in the area of disease and pest detection. Recent research works in pest and insect detection \cite{pest1, pest2, pest3, pest4, pest5}, invasive species detection in marine aquaculture \cite{aquaculture} and disease detection in plant leafs \cite{leaf-disease1, leaf-disease2, leaf-disease3, leaf-disease4, leaf-disease5}, Rice \cite{rice1, rice2, rice3}, Tomato \cite{pest-recog-tomato, tomato, tomato2, tomato3}, Banana \cite{banana}, Grape \cite{grape}, Sugarcane \cite{sugarcane}, Eggplant \cite{eggplant}, Cucumber \cite{cucumber}, Soybean \cite{soybean}, Olive \cite{olive}, Tea \cite{tea}, Coffee \cite{coffee} and other similar works take encouraging steps towards disease-free agriculture. Fig. \ref{fig-disease} depicts banana diseases and pest detection outputs from \cite{banana}. This work \cite{disease-detection-limits} reports solutions to extant limitations in plant disease detection.  

\subsection{Unmanned Aircraft Vehicles for Plant Phenotyping}
The past few decades have witnessed the great progress of unmanned aircraft vehicles (UAVs) in civilian fields, especially in photogrammetry and remote sensing. In contrast with the platforms of manned aircraft and satellite, the UAV platform holds many promising characteristics: flexibility, efficiency, high spatial/temporal resolution, low cost, easy operation, etc., which make it an effective complement to other remote-sensing platforms and a cost-effective means for remote sensing. We refer reader to literary works \cite{uav-survey1, uav-survey2} for the detailed reports of techniques and applications of UAVs in precision agriculture, remote sensing, search and rescue, construction and infrastructure inspection and discuss other market opportunities. UAVs can be utilized in precision agriculture (PA) for crop management and monitoring \cite{uav-cropmanagement, uav-cropmonitoring}, weed detection \cite{uav-weed-detect}, irrigation scheduling \cite{uav-irrigation}, agricultural pattern detection \cite{uav-agrovision}, pesticide spraying \cite{uav-cropmanagement}, cattle detection \cite{uav-cattle}, disease detection \cite{uav-disease, uav-disease2}, insect detection \cite{uav-insect} and data collection from ground sensors (moisture, soil properties, etc.,) \cite{uav-datacollect}. The deployment of UAVs in PA is a cost-effective and time saving technology which can help for improving crop yields, farms productivity and profitability in farming systems. Moreover, UAVs facilitate agricultural management, weed monitoring, and pest damage, thereby they help to meet these challenges quickly \cite{uav-agromanagement}. 

UAVs can also be utilized to monitor and quantify several factors of irrigation such as availability of soil water, crop water need (which represents the amount of water needed by the various crops to grow optimally), rainfall amount, efficiency of the irrigation system \cite{uav-irrigation-soil}. In this work \cite{uav-soilpaper}, UAVs are currently being utilized to estimate the spatial distribution of surface soil moisture high-resolution multi-spectral imagery in combination with ground sampling. UAVs are also being used for thermal remote sensing to monitor the spatial and temporal patterns of crop diseases during various disease development phases which reduces crop losses for farmers. This work \cite{uav-soil-fungus} detects early stage development of soil-borne fungus in UAV imagery.  Soil texture can be an indicative of soil quality which in turn influences crop productivity. Hence, UAV thermal images are being utilized to quantify soil texture at a regional scale by measuring the differences in land surface temperature under a relatively homogeneous climatic condition \cite{uav-soiltexture1, uav-soiltexture2}. Accurate assessment of crop residue is crucial for proper implementation of conservation tillage practices since crop residues provide a protective layer on agricultural fields that shields soil from wind and water. In \cite{uav-soilresidue}, the authors demonstrated that aerial thermal images can explain more than 95\% of the variability in crop residue cover amount compared to 77\% using visible and near IR images.

Farmers must monitor crop maturity to determine the harvesting time of their crops. UAVs can be a practical solution to this problem \cite{uav-crop-maturity}. Farmers require accurate, early estimation of crop yield for a number of reasons, including crop insurance, planning of harvest and storage requirements, and cash flow budgeting. In \cite{uav-crop-yield1}, UAV images were utilized to estimate yield and total biomass of rice crop in Thailand. In \cite{uav-crop-yield2}, UAV images were also utilized to predict corn grain yields in the early to midseason crop growth stages in Germany. 

There have also been successful efforts that seamlessly combine aerial and ground based system for precision agriculture \cite{uav-ground}. With relaxed flight regulations and drastic improvement in machine learning techniques, geo-referencing, mosaicing, and other related algorithms, UAVs can provide a great potential for soil and crop monitoring \cite{uav-conclusion}. More precision agricultural researches are encouraged to design and implement special types of cameras and sensors on-board UAVs, which have the ability of remote crop monitoring and detection of soil and other agricultural characteristics in real time scenarios.

\subsection{Satellites for Plant Phenotyping}

The impact of climate change and its unforeseeable nature, has caused majority of the agricultural crops to be affected in terms of their production and maintenance. With more than seven billion mouths to feed greater demands are being put on agriculture than ever before, at the same time as land is being degraded by factors such as soil erosion, mineral exhaustion and drought. It becomes the utmost priority for governments to support farmers by providing crucial information about changing weather conditions, soil conditions and more. Currently, satellite imagery is making agriculture more efficient by reducing scouting efforts of farmers, by optimizing use of nitrogen based on variable rate of application, by optimizing water schedules, identifying field performance and benchmark fields, etc \cite{satellites-gamaya}. India alone has 7 satellites specially designed for benefits of farmers \cite{satellites-india}.

Satellites and their imagery are being applied to agriculture in several ways, initially as a means of estimating crop yields \cite{esa-earth-online2} and crop types \cite{satellite-crop-type}, soil salinity, soil moisture, soil pH  \cite{satellite-soil3, satellite-soil1, satellite-soil2}. Optical and radar sensors can provide an accurate picture of the acreage being cultivated, while also differentiating between crop types and determining their health and maturity. Optical satellite sensors can detect visible and near-infrared wavelengths of light, reflected from agricultural land below. It is these wavelengths which combined, can be manipulated to help us understand the condition of the crops. This information helps to inform the market, and provide early warning of crop failure or famine.

By extension, satellites are also used as a management tool through the practice of PA, where satellite images are used to characterise a farmer's fields in detail, often used in combination with geographical information systems (GIS), to allow more intensive and efficient cultivation practices. For instance, different crops might be recommended for different fields while the farmer's use of fertiliser is optimised in a more economic and environmentally-friendly fashion. Providing access to satellite imagery also becomes very important for building trust among the involved parties (farmers and government and private bodies involved). Web-based platforms such as Google Earth Engine, Planet.com, Earth Data Search by NASA, LandViewer by Earth Observing System, Geocento \cite{geocento} and others \cite{satellite-imagery-sources} provide access to past and present (even daily) satellite imagery of your interest. 

Agricultural monitoring is also increasingly being applied to forestry, both for forest management and as a way of characterising forests as carbon sinks to help minimise climate change – notably as part of the UN's REDD programme \cite{esa-earth-online}.

\section{Plant Phenotyping with Limited Labeled Data}

While deep learning based plant phenotyping has shown great promise, requirement of large labeled datasets still remains to be the bottleneck. Phenotyping tasks are often specific to the environmental and genetic conditions, finding large datasets with such conditions is not always possible. This results in researchers needing to acquire their own dataset and label it, which is often a arduous and expensive affair. Moreover, small datasets often lead to models that overfit. Deep learning approaches optimized for working with limited labeled data would immensely help the plant phenotyping community, since this would encourage many more farmers, breeders, and researchers to employ reliable plant phenotyping techniques to optimize crop yield. To this end, we list out some of the recent efforts in the area of deep plant phenotyping with limited labeled data.

\subsection*{Data Augmentation}
\begin{figure}
\centerline{\includegraphics[scale=0.55]{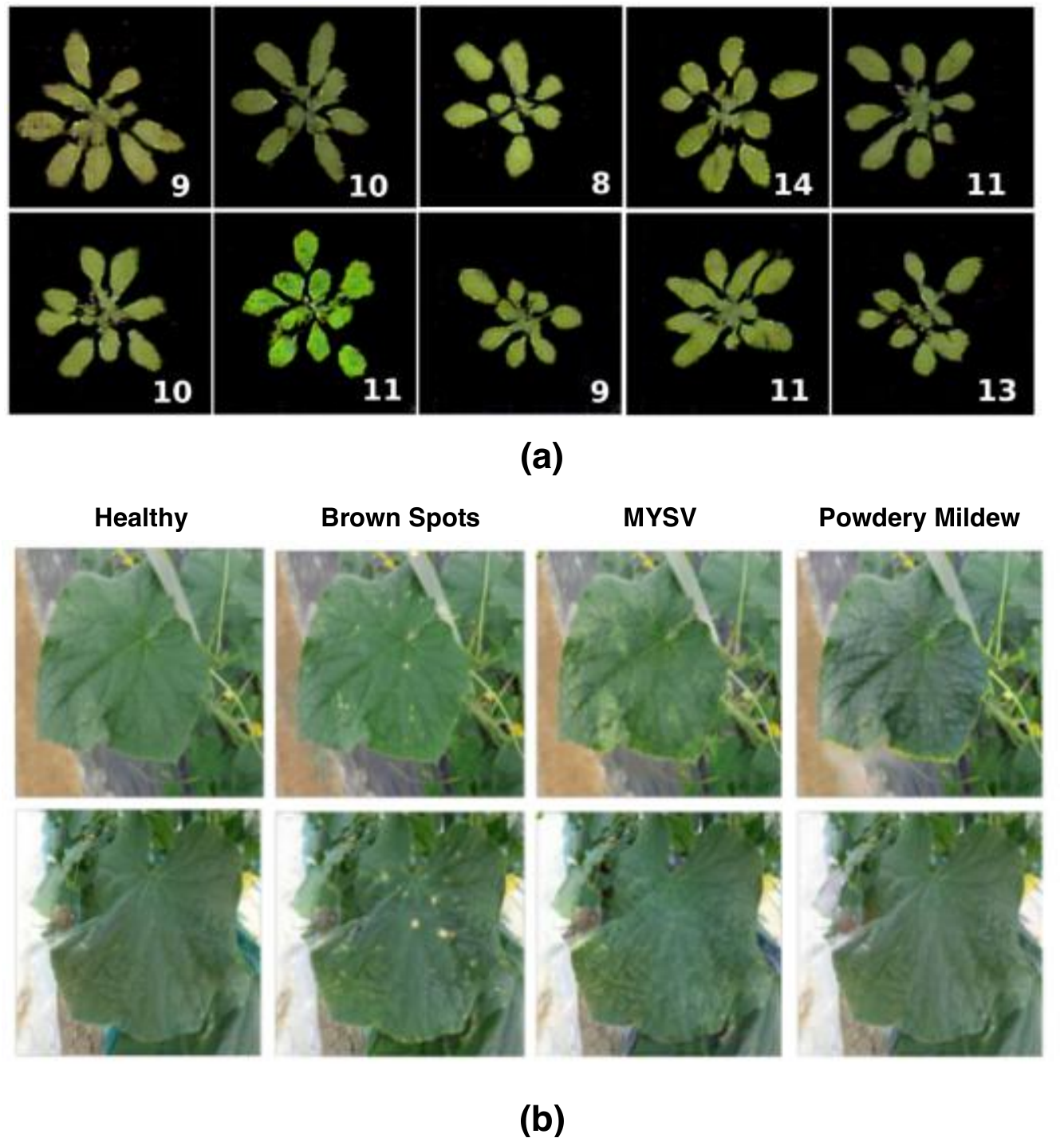}}
\caption{\textbf{(a)} shows \textit{Arabidopsis} plant images generated by ARIGAN \cite{data-aug-arigan}. Bottom-right numbers refer to the leaf count. \textbf{(b)} shows two sets of healthy leafs and their corresponding disease prone leaves generated by LeafGAN \cite{leafgan}.}
\label{fig-generated}
\end{figure}

The computer vision community has long been employing dataset augmentation techniques to grow the amount of data using artificial transformations. Artificially perturbing the original dataset with affine transformations (e.g., rotation, scale, translation) is considered a common practice now. However, this approach has some constraints: the augmented data only capture the variability of the available training set (e.g., if the dataset doesn't include a unique colored fruit, the particular unique case will never be learnt). To overcome this, several data augmentation methods proposed take advantage of recent advancements in the image generation space. In this work \cite{data-aug-arigan}, the authors use Generative Adversarial Network (GAN) \cite{gans} to generate \textit{Arabidopsis} plant images (called ARIGAN) with unique desirable traits (over 7 leaves) that were originally less frequent in the dataset. Fig. \ref{fig-generated} \textbf{(a)} shows examples of images generated by ARIGAN. Other latest works \cite{data-aug-rosette1, data-aug-rosette2} use more advanced variants of GANs to generate realistic plant images with particularly favorable leaf segmentations of interest to boost leaf counting accuracy of the learning models. In \cite{data-aug-unsup-gan}, the authors proposed an unsupervised image translation technique to improve plant disease recognition performance. LeafGAN \cite{leafgan}, an image-to-image translation model, generates leaf images with various plant diseases and boosts diagnostic performance by a great margin. Two sets of example images generated by LeafGAN are shown in Fig. \ref{fig-generated} \textbf{(b)}. Other data enhancement techniques are also being employed by researchers to train plant disease diagnosis models on generated lesions \cite{data-enhance-disease}. 

The effort to provide finely annotated data has enabled great improvement of the state of the art on segmentation performance. Some researches have started working on effectively transferring the knowledge obtained from RGB images on annotated plants either to other species or other modalities of imaging. In this work \cite{data-aug-miltimodal}, the authors successfully transfer the knowledge gained from annotated leaves of \textit{Arabidopsis thaliana} in RGB to images of the same plant in chlorophyll fluorescence imaging. 

\subsection*{Weakly Supervised Learning}
Fruit/organ counting is a well explored task by the plant phenotyping community. However, many vision-based solutions we have currently require highly accurate instance and density labels of fruits and organs in diverse set of environments. The labeling procedures are often very burdensome and error prone and, in many agricultural scenarios, it may be impossible to acquire a sufficient number of labelled samples to achieve consistent performance that are robust to image noise or other forms of covariate shift. This is why using only weak labels can be crucial for cost-effective plant phenotyping. 

Recently, a lot of attention has been placed on engineering weakly supervised learning frameworks for plant phenotyping. In \cite{weak-sup1}, the authors created a weakly supervised framework for the sorghum head detection task where annotators label the data only until the model reaches a desired performance level. After that, model outputs are directly passed as data labels leading to a exponential reduction in annotation costs with minimal loss in model accuracy. In other work \cite{weak-sup2}, the authors proposed a strategy which is able to learn to count fruits without requiring task-specific supervision labels, such as manually labelled object bounding boxes or total instance count. In \cite{weak-sup3}, the authors use a trained CNN on defect classification data and use it's activate maps to segment infected regions on potatoes. Segmentation task requires really rich labels (each pixel of the image is annotated) so this task effectively bypasses the labeling for segmentation altogether. On another note, rice heading date estimation greatly assists the breeders to understand the adaptability of the crop to various environmental and genetic conditions. Accurate estimation of heading date requires monitoring the increase in number of rice panicles in the crop. Detecting rice panicles from crop images usually requires training an object detection model such as Faster R-CNN or YOLO, which requires costly bounding box annotations. However, a recently proposed method \cite{Desai2019} uses a sliding window based detector which requires training an image classifier, for which annotations are much easier to obtain. 
\begin{figure}
\centerline{\includegraphics[scale=0.3]{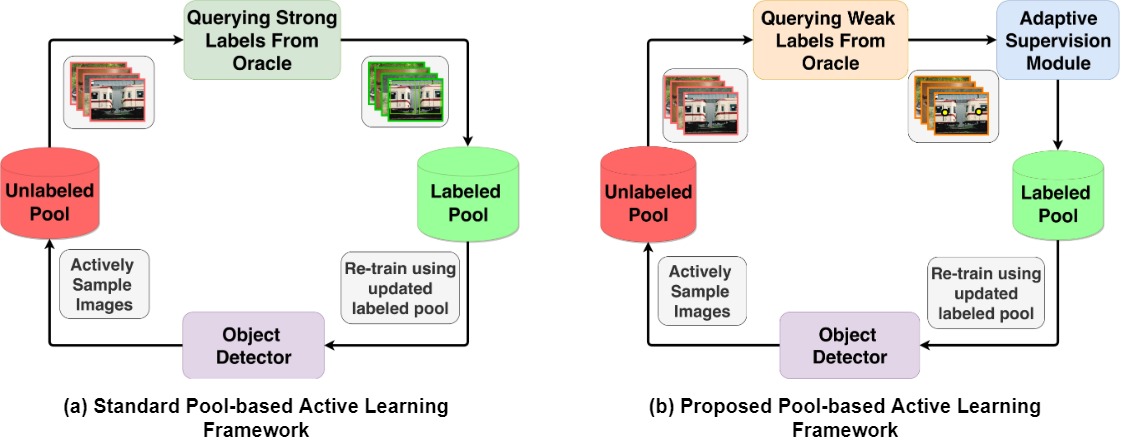}}
\caption{\textbf{(a)} Standard pool-based active learning framework \textbf{(b)} Proposed framework \cite{Desai2019AnAS} which interleaves weak supervision in the active learning process. This novel framework includes an adaptive supervision module which allows switching to a stronger form of supervision as required when training the model. The \textit{oracle} is the source of labels \textit{a.k.a} annotator.}
\label{fig-bmvc}
\end{figure}

\subsection*{Transfer Learning}
Transfer learning is a type of learning that enables using knowledge gained while solving one problem and applying it to a different but related problem i.e., a model trained on one phenotyping task (say potato leaf classification) being able to assist another phenotyping (tomato leaf classification) task. Transfer learning is a very well explored area of machine learning. As part of the first steps of adopting existing transfer learning techniques for plant phenotyping, the authors of \cite{transfer-learning-2} use CNNs (AlexNet, GoogleNet and VGGNet) pretrained on ImageNet dataset \cite{imagenet} and fine tune on the plant dataset used in LifeCLEF \cite{lifeclef} 2015 challenge. With the help of transfer learning, they were able to beat then existing state-of-the-art LifeCLEF performance by 15\% points. Similary in \cite{transfer-learning-1}, the authors report better than human results in segmentation task with the help of transfer learning where they transfer learn a model trained on peanut root dataset for switchgrass root dataset (they also report results using ImageNet pretrained models). Leaf disease detection and treatment recommendation performance is also shown to be boosted with transfer learning \cite{tl-leaf-disease}.  In \cite{domain-adapt}, the authors interestingly combined a State-of-the-Art weakly-supervised fruit counting model with an unsupervised style transfer method for fruit counting. They used Cycle-Generative Adversarial Network (C-GAN) to perform unsupervised domain adaptation from one fruit dataset to another and train it alongside with a Presence-Absence Classifier (PAC) that discriminates images containing fruits or not and ultimately achieved better performance than fully supervised models.

\begin{figure}
\centerline{\includegraphics[scale=0.15]{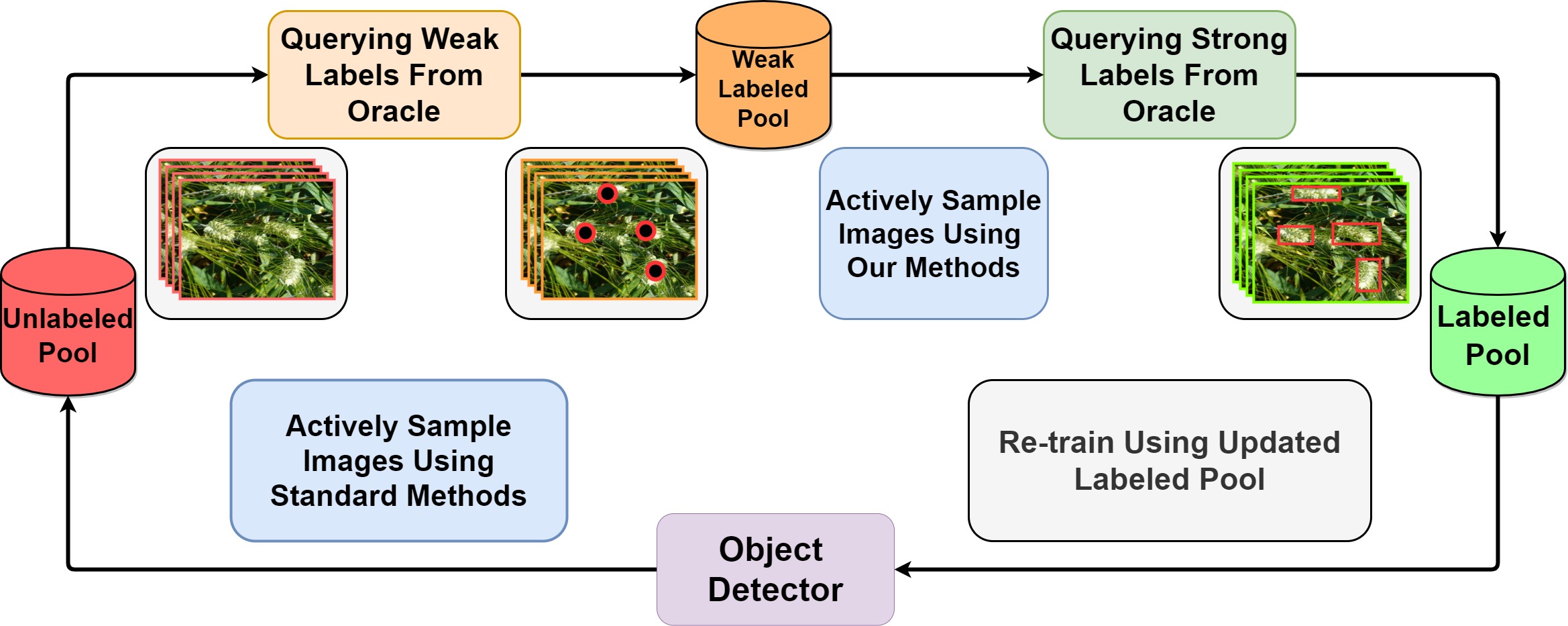}}
\caption{Proposed point supervision framework \cite{ch2019active} into the pool-based active learning cycle. In this framework, strong supervision is queried for images only after deemed \textit{informative} based on point supervision of those images.}
\label{fig-bmc}
\end{figure}

\subsection*{Active Learning}
Active learning \cite{Settles10activelearning}, an iterative training approach that curiously selects the best samples to train, has been shown to reduce labeled data requirement when training deep classification networks \cite{Gal2017DeepBA, Sener2018ActiveLF, Wang_gupta}. Research in the area of active learning for object detection \cite{al-deep-object-baseline, VinayNamboodiri2018, Vijayanarasimhan2014} has been, arguably, limited. However, numerous plant phenotyping tasks such as detection and quantification of crop yield and fruit counting are directly dependent on object detection. Keeping this in mind, an active learning method has been proposed \cite{Desai2019AnAS} for training deep object detection models where the model can selectively query either weak labels (pointing at the object) or strong labels (drawing a box around the object). By introducing a switching module for weak labels and strong labels, the authors were able to save 24\% of annotation time while training a wheat head detection \cite{wheat-madec} model. Fig. \ref{fig-bmvc} illustrates the difference between regular active learning cycle and proposed active learning cycle. This method demonstrates the applicability of active learning to plant phenotyping methods where obtaining labeled data is often difficult. Along the same lines, to alleviate the labeled data requirement for training object detection models for cereal crop detection, a weak supervision based active learning method \cite{ch2019active} was proposed recently. In this active learning approach, the model constantly interacts with a human annotator by iteratively querying the labels for only the most informative images, as opposed to all images in a dataset. Fig. \ref{fig-bmc} visually illustrates the proposed framework. The active query method is specifically designed for cereal crops which usually tend to have panicles with low variance in appearance. This training method has been shown to reduce over 50\% of annotation costs on sorghum head and wheat spike detection datasets. We expect to see more research works using active learning for limited labeled data based plant phenotyping in the near future.

\section{Challenges and Open Problems}

In this section, we describe some of the challenges present in plant phenotyping methods which warrant further research. 

\subsection*{The Training Data Bottleneck} 

Modern phenotyping methods rely on deep learning which is notorious for requiring large amounts of labeled data. While some progress has been made in developing data efficient models for phenotyping, reducing the labeling efforts for training efficient phenotyping tools is still an open problem. We believe that effectively adapting techniques from deep learning such as unsupervised, self supervised, weakly supervised, active and semi-supervised learning will greatly benefit the phenotyping community in observing plant traits with small datasets. 

\subsection*{Explainability}

Deep neural networks are generally considered as black boxes which produce predictions without sufficient justification. This makes debugging a neural network difficult i.e., it can be tough to understand what caused a wrong prediction. Crop management decisions based on incorrect phenotyping results can cause financial losses. Hence, developing explainable models for plant phenotyping is one of the open problems in this field. Obtaining the reasons behind a given set of plant traits using explainable models has the potential to achieve breakthroughs in our understanding of plant behavior in various genetic and environmental conditions. 

\subsection*{Data collection}

Vision based plant phenotyping suffers from challenges such as occlusion, inaccuracies in 3D reconstruction of crops and bad lighting conditions caused by the changing weather. It is therefore necessary to develop phenotyping tools which are robust to visual variations.

\section{Conclusions}

High throughput plant phenotyping methods have shown great promise in efficiently monitoring crops for plant breeding and agricultural crop management. Research in deep learning has accelerated the progress in plant phenotyping research which resulted in the development of various image analysis tools to observe plant traits. However, wide applicability of high throughput phenotyping tools is limited by some issues such as 1) dependence of deep networks on large datasets, which are difficult to curate, 2) large variations of field environment which cannot always be captured, and 3) capital and maintenance which can be prohibitively expensive to be widely used in developing countries. With many open problems in plant phenotyping warranting further studies, it is indeed a great time to study plant phenotyping and achieve rapid progress by utilizing the advances in deep learning.   


\bibliographystyle{splncs}
\bibliography{egbib}

\begin{thebibliography}{100}

\bibitem{Agro2050}
Hunter, M., Smith, R., Schipanski, M., Atwood, L., Mortensen, D.:
\newblock Agriculture in 2050: Recalibrating targets for sustainable
  intensification.
\newblock BioScience \textbf{67} (02 2017)

\bibitem{arable}
Wu, X., Guo, J., Han, M., Chen, G.:
\newblock An overview of arable land use for the world economy: From source to
  sink via the global supply chain.
\newblock Land Use Policy \textbf{76} (2018)  201 -- 214

\bibitem{sust1}
Gago, J., Douthe, C., Coopman, R., Gallego, P., Ribas-Carbo, M., Flexas, J.,
  Escalona, J., Medrano, H.:
\newblock Uavs challenge to assess water stress for sustainable agriculture.
\newblock Agricultural Water Management \textbf{153} (2015)  9 -- 19

\bibitem{sust2}
Aubert, B.A., Schroeder, A., Grimaudo, J.:
\newblock It as enabler of sustainable farming: An empirical analysis of
  farmers' adoption decision of precision agriculture technology.
\newblock Decision Support Systems \textbf{54}(1) (2012)  510 -- 520

\bibitem{sust3}
Sladojevic, S., Arsenovic, M., Anderla, A., Culibrk, D., Stefanovic, D.:
\newblock Deep neural networks based recognition of plant diseases by leaf
  image classification.
\newblock In: Comp. Int. and Neurosc. (2016)

\bibitem{bonirob}
Ruckelshausen, A., Biber, P., Dorna, M., Gremmes, H., Klose, R., Linz, A.,
  Rahe, R., Resch, R., Thiel, M., Trautz, D., Weiss, U.:
\newblock Bonirob: An autonomous field robot platform for individual plant
  phenotyping.
\newblock Precision Agriculture \textbf{9} (01 2009)  841--847

\bibitem{fitch_1944}
Fitch, F.B.:
\newblock Warren s. mcculloch and walter pitts. a logical calculus of the ideas
  immanent in nervous activity. bulletin of mathematical biophysics, vol. 5
  (1943), pp. 115–133.
\newblock Journal of Symbolic Logic \textbf{9}(2) (1944)  49–50

\bibitem{Rosenblatt1958ThePA}
Rosenblatt, F.F.:
\newblock The perceptron: a probabilistic model for information storage and
  organization in the brain.
\newblock Psychological review \textbf{65 6} (1958)  386--408

\bibitem{universalapprox}
Hornik, K., Stinchcombe, M., White, H.:
\newblock Multilayer feedforward networks are universal approximators.
\newblock Neural Netw. \textbf{2}(5) (July 1989)  359–366

\bibitem{unsuppretraining}
Bengio, Y., Lamblin, P., Popovici, D., Larochelle, H., Montreal, U.:
\newblock Greedy layer-wise training of deep networks.
\newblock Volume~19. (01 2007)

\bibitem{taskagnosticbengio}
Bengio, Y.:
\newblock Learning deep architectures for ai.
\newblock Foundations \textbf{2} (01 2009)  1--55

\bibitem{Alom_2019}
Alom, M.Z., Taha, T.M., Yakopcic, C., Westberg, S., Sidike, P., Nasrin, M.S.,
  Hasan, M., Van~Essen, B.C., Awwal, A.A.S., Asari, V.K.:
\newblock A state-of-the-art survey on deep learning theory and architectures.
\newblock Electronics \textbf{8}(3) (Mar 2019)  292

\bibitem{GAN}
Goodfellow, I., Pouget-Abadie, J., Mirza, M., Xu, B., Warde-Farley, D., Ozair,
  S., Courville, A., Bengio, Y.:
\newblock Generative adversarial nets.
\newblock In Ghahramani, Z., Welling, M., Cortes, C., Lawrence, N.D.,
  Weinberger, K.Q., eds.: Advances in Neural Information Processing Systems 27.
\newblock Curran Associates, Inc. (2014)  2672--2680

\bibitem{cnn-figure-ref}
:
\newblock A framework for designing the architectures of deep convolutional
  neural networks.
\newblock Entropy \textbf{19}(6) (May 2017)  242

\bibitem{Fukushima1988NeocognitronAH}
Fukushima, K.:
\newblock Neocognitron: A hierarchical neural network capable of visual pattern
  recognition.
\newblock Neural Networks \textbf{1} (1988)  119--130

\bibitem{Lecun98Gradient}
{Lecun}, Y., {Bottou}, L., {Bengio}, Y., {Haffner}, P.:
\newblock Gradient-based learning applied to document recognition.
\newblock Proceedings of the IEEE \textbf{86}(11) (Nov 1998)  2278--2324

\bibitem{fcnn-cnn-figure}
Itay~Lieder, Yehezkel S.~Resheff, T.H.:
\newblock Learning tensorflow.
\newblock (2017)  Chapter 4

\bibitem{pooling-paper}
Boureau, Y.L., Ponce, J., Lecun, Y.:
\newblock A theoretical analysis of feature pooling in visual recognition.
\newblock (11 2010)  111--118

\bibitem{pooling-paper2}
Scherer, D., M{\"u}ller, A., Behnke, S.:
\newblock Evaluation of pooling operations in convolutional architectures for
  object recognition.
\newblock In Diamantaras, K., Duch, W., Iliadis, L.S., eds.: Artificial Neural
  Networks -- ICANN 2010, Berlin, Heidelberg, Springer Berlin Heidelberg (2010)
   92--101

\bibitem{det-survey}
Jiao, L., Zhang, F., Liu, F., Yang, S., Li, L., Feng, Z., Qu, R.:
\newblock A survey of deep learning-based object detection.
\newblock IEEE Access \textbf{7} (2019)  128837--128868

\bibitem{seg-survey}
Lu, Z., Xu, H., Liu, G.:
\newblock A survey of object co-segmentation.
\newblock IEEE Access \textbf{PP} (05 2019)  1--1

\bibitem{mscoco}
Lin, T.Y., Maire, M., Belongie, S.J., Hays, J., Perona, P., Ramanan, D.,
  Doll{\'a}r, P., Zitnick, C.L.:
\newblock Microsoft coco: Common objects in context.
\newblock In: ECCV. (2014)

\bibitem{cropweed4}
Y.~Gharde, P.~Singh, R.D., Gupta, P.:
\newblock Assessment of yield and economic losses in agriculture due to weeds
  in india.
\newblock Volume 107. (2018)  12--18

\bibitem{cropweed}
Haug, S., Ostermann, J.:
\newblock A crop/weed field image dataset for the evaluation of computer vision
  based precision agriculture tasks.
\newblock In: ECCV Workshops. (2014)

\bibitem{cropweed2}
Binguitcha-Fare, A.A., Sharma, P.:
\newblock Crops and weeds classification using convolutional neural networks
  via optimization of transfer learning parameters

\bibitem{cropweed3}
Fawakherji, M., Youssef, A., Bloisi, D.D., Pretto, A., Nardi, D.:
\newblock Crop and weeds classification for precision agriculture using
  context-independent pixel-wise segmentation.
\newblock 2019 Third IEEE International Conference on Robotic Computing (IRC)
  (2019)  146--152

\bibitem{cropweed5}
Guerrero, J.M., Pajares, G., Montalvo, M., Romeo, J., Guijarro, M.:
\newblock Support vector machines for crop/weeds identification in maize
  fields.
\newblock Expert Syst. Appl. \textbf{39} (2012)  11149--11155

\bibitem{weed-net}
Sa, I., Chen, Z., Popovic, M., Khanna, R., Liebisch, F., Nieto, J., Siegwart,
  R.:
\newblock weednet: Dense semantic weed classification using multispectral
  images and mav for smart farming.
\newblock IEEE Robotics and Automation Letters \textbf{3} (2017)  588--595

\bibitem{weed-id}
Rani, K., Supriya, P., Sarath, T.V.:
\newblock Computer vision based segregation of carrot and curry leaf plants
  with weed identification in carrot field.
\newblock 2017 International Conference on Computing Methodologies and
  Communication (ICCMC) (2017)  185--188

\bibitem{plant-seedlings1}
Nkemelu, D.K., Omeiza, D., Lubalo, N.:
\newblock Deep convolutional neural network for plant seedlings classification.
\newblock ArXiv \textbf{abs/1811.08404} (2018)

\bibitem{plant-seedlings2}
Elnemr, H.A.:
\newblock Convolutional neural network architecture for plant seedling
  classification.
\newblock (2019)

\bibitem{miniUAV}
Xiang, T.Z., Xia, G.S., Zhang, L.:
\newblock Mini-uav-based remote sensing: Techniques, applications and
  prospectives (12 2018)

\bibitem{crop-categorization}
Patr{\'i}cio, D.I., Rieder, R.:
\newblock Computer vision and artificial intelligence in precision agriculture
  for grain crops: A systematic review.
\newblock Comput. Electron. Agric. \textbf{153} (2018)  69--81

\bibitem{pest-recog-tomato}
Fuentes, A., Yoon, S., Kim, S.C., Park, D.S.:
\newblock A robust deep-learning-based detector for real-time tomato plant
  diseases and pests recognition.
\newblock In: Sensors. (2017)

\bibitem{harvesting-systems}
Bachche, S.:
\newblock Deliberation on design strategies of automatic harvesting systems: A
  survey.
\newblock Robotics \textbf{4} (2015)  194--222

\bibitem{irrigation-growth-monitor}
Allende, A., Monaghan, J.M., Uyttendaele, M., Franz, E., Schl{\"u}ter, O.:
\newblock Irrigation water quality for leafy crops: A perspective of risks and
  potential solutions.
\newblock In: International journal of environmental research and public
  health. (2015)

\bibitem{sorghum-guo}
Guo, W., Zheng, B., Potgieter, A.B., Diot, J., Watanabe, K., Noshita, K.,
  Jordan, D.R., Wang, X., Watson, J., Ninomiya, S., Chapman, S.C.:
\newblock {Aerial Imagery Analysis – Quantifying Appearance and Number of
  Sorghum Heads for Applications in Breeding and Agronomy}.
\newblock Frontiers in Plant Science \textbf{9}(October) (2018)  1--9

\bibitem{irrigation}
Chai, Q., Gan, Y., Zhao, C., Xu, H.l., Waskom, R., Niu, Y., Siddique, K.:
\newblock Regulated deficit irrigation for crop production under drought
  stress. a review.
\newblock Agronomy for Sustainable Development \textbf{36} (03 2016)

\bibitem{apple-picking}
Zhao, D., Liu, X., Chen, Y., Ji, W., Jia, W., Hu, C.:
\newblock Image recognition at night for apple picking robot.
\newblock Nongye Jixie Xuebao/Transactions of the Chinese Society for
  Agricultural Machinery \textbf{46} (03 2015)  15--22

\bibitem{pesticide-spraying}
Yamane, S., Miyazaki, M.:
\newblock Study on electrostatic pesticide spraying system for
  low-concentration, high-volume applications.
\newblock (2017)

\bibitem{fertilization}
Oktay, K., Bedoschi, G., Pacheco, F., Turan, V., Emirdar, V.:
\newblock First pregnancies, livebirth and in vitro fertilization outcomes
  after transplantation of frozen-banked ovarian tissue with a human
  extracellular matrix scaffold using robot-assisted minimally invasive
  surgery.
\newblock American Journal of Obstetrics and Gynecology \textbf{214} (11 2015)

\bibitem{cropdeep}
Zheng, Y.Y., Kong, J.L., bo~Jin, X., Wang, X.Y., Su, T.L., Zuo, M.:
\newblock Cropdeep: The crop vision dataset for deep-learning-based
  classification and detection in precision agriculture.
\newblock In: Sensors. (2019)

\bibitem{rosette1}
Minervini, M., Fischbach, A., Scharr, H., Tsaftaris, S.:
\newblock Finely-grained annotated datasets for image-based plant phenotyping.
\newblock Pattern Recognition Letters \textbf{81} (11 2015)

\bibitem{rosette2}
Scharr, H., Minervini, M., Fischbach, A., Tsaftaris, S.:
\newblock Annotated image datasets of rosette plants (07 2014)

\bibitem{pp-dataset}
Cruz, J., Yin, X., Liu, X., Imran, S., Morris, D., Kramer, D., Chen, J.:
\newblock Multi-modality imagery database for plant phenotyping.
\newblock Machine Vision and Applications \textbf{27} (07 2016)

\bibitem{wheat-madec}
Madec, S., Jin, X., Lu, H., de~Solan, B., Liu, S., Duyme, F., Heritier, E.,
  Frederic, B.:
\newblock Ear density estimation from high resolution rgb imagery using deep
  learning technique.
\newblock Agricultural and Forest Meteorology \textbf{264} (01 2019)  225--234

\bibitem{crop-tassel-dataset}
Lu, H., Cao, Z.G., Xiao, Y., Li, Y., Zhu, Y.:
\newblock Joint crop and tassel segmentation in the wild.
\newblock (11 2015)

\bibitem{weak-sup1}
Ghosal, S., Zheng, B., Chapman, S.C., Potgieter, A.B., Jordan, D., Wang, X.,
  Singh, A.K., Singh, A., Hirafuji, M., Ninomiya, S., Ganapathysubramanian, B.,
  Sarkar, S., Guo, W.:
\newblock A weakly supervised deep learning framework for sorghum head
  detection and counting.
\newblock (2019)

\bibitem{Desai2019}
Desai, S.V., Balasubramanian, V.N., Fukatsu, T., Ninomiya, S., Guo, W.:
\newblock {Automatic estimation of heading date of paddy rice using deep
  learning}.
\newblock Plant Methods \textbf{15}(1) (2019) ~76

\bibitem{wspike}
Hasan, M.M., Chopin, J.P., Laga, H., Miklavcic, S.J.:
\newblock Detection and analysis of wheat spikes using convolutional neural
  networks.
\newblock Plant Methods \textbf{14}(1) (Nov 2018)  100

\bibitem{leaf_counting}
Ubbens, J., Cieslak, M., Prusinkiewicz, P., Stavness, I.:
\newblock The use of plant models in deep learning: an application to leaf
  counting in rosette plants.
\newblock In: Plant Methods. (2018)

\bibitem{Sa2016DeepFruitsAF}
Sa, I., Ge, Z., Dayoub, F., Upcroft, B., Perez, T., McCool, C.:
\newblock Deepfruits: A fruit detection system using deep neural networks.
\newblock In: Sensors. (2016)

\bibitem{panicle_seg}
Xiong, X., Duan, L., Liu, L., Tu, H., Yang, P., Wu, D., Chen, G., Xiong, L.,
  Yang, W., Liu, Q.:
\newblock Panicle-seg: a robust image segmentation method for rice panicles in
  the field based on deep learning and superpixel optimization.
\newblock Plant Methods \textbf{13}(1) (Nov 2017)  104

\bibitem{Oh2019}
Oh, M.h., Olsen, P., Ramamurthy, K.N.:
\newblock {Counting and Segmenting Sorghum Heads}.
\newblock (may 2019)

\bibitem{food-security}
Tai, A., Val~Martin, M., Heald, C.:
\newblock Threat to future global food security from climate change and ozone
  air pollution.
\newblock Nature Climate Change \textbf{4} (07 2014)  817--821

\bibitem{pollinators}
Unknown:
\newblock Pollinators vital to our food supply under threat.
\newblock Volume Press Release., Intergovernmental Platform on Biodiversity and
  Ecosystem Services (2016)

\bibitem{plant-disease}
Strange, R., Scott, P.:
\newblock Plant disease: A threat to global food security.
\newblock Annual review of phytopathology \textbf{43} (02 2005)  83--116

\bibitem{plantdoc}
Singh, D., Jain, N., Jain, P., Kayal, P., Kumawat, S., Batra, N.:
\newblock Plantdoc: A dataset for visual plant disease detection.
\newblock In: CoDS COMAD 2020. (2020)

\bibitem{smallholders}
Walpole, M., Smith, J., Rosser, A., Brown, C., Schulte-Herbruggen, B., Booth,
  H., Sassen, M., Mapendembe, A., Fancourt, M., Bieri, M., Glaser, S.,
  Corrigan, C., Narloch, U., Runsten, L., Jenkins, M., Gomera, M., Hutton, J.:
\newblock Smallholders, food security, and the environment (03 2013)

\bibitem{madagascar}
Harvey Celia~A., Rakotobe Zo~Lalaina, R.N.S.D.R.R.H.R.R.H.R.H., L., M.J.:
\newblock Extreme vulnerability of smallholder farmers to agricultural risks
  and climate change in madagascar (04 2014)

\bibitem{world-hunger}
Sanchez, P., Swaminathan, M.:
\newblock Cutting world hunger in half.
\newblock Science (New York, N.Y.) \textbf{307} (02 2005)  357--9

\bibitem{kisan-helpline}
MinistryOfAgriculture:
\newblock Government of india 2019. kisaan knowledge management system.
\newblock (2019)

\bibitem{plant-village}
Hughes, D., Salathe, M.:
\newblock An open access repository of images on plant health to enable the
  development of mobile disease diagnostics through machine learning and
  crowdsourcing.
\newblock (11 2015)

\bibitem{pest1}
Liu, B., Zhuhua, H., Zhao, Y., Bai, Y., Wang, Y.:
\newblock Recognition of pyralidae insects using intelligent monitoring
  autonomous robot vehicle in natural farm scene (01 2019)

\bibitem{pest2}
Deng, L., Wang, Y., Han, Z., Yu, R.:
\newblock Research on insect pest image detection and recognition based on
  bio-inspired methods.
\newblock (2018)

\bibitem{pest3}
Javed, M.H., Humair, M., Yaqoob, B., Noor, N., Arshad, T.:
\newblock K-means based automatic pests detection and classification for
  pesticides spraying.
\newblock International Journal of Advanced Computer Science and Applications
  \textbf{8} (01 2017)

\bibitem{pest4}
Liu, T., Chen, W., Wu, W., Sun, C., Guo, W., Zhu, X.:
\newblock Detection of aphids in wheat fields using a computer vision
  technique.
\newblock Biosystems Engineering \textbf{141} (01 2016)  82--93

\bibitem{pest5}
Zhong, Y., Gao, J., Lei, Q., Zhou, Y.:
\newblock A vision-based counting and recognition system for flying insects in
  intelligent agriculture.
\newblock In: Sensors. (2018)

\bibitem{aquaculture}
Galloway, A., Taylor, G.W., Ramsay, A., Moussa, M.A.:
\newblock The ciona17 dataset for semantic segmentation of invasive species in
  a marine aquaculture environment.
\newblock 2017 14th Conference on Computer and Robot Vision (CRV) (2017)
  361--366

\bibitem{leaf-disease1}
Zhang, S., Wu, X., You, Z.H., Zhang, L.:
\newblock Leaf image based cucumber disease recognition using sparse
  representation classification.
\newblock Comput. Electron. Agric. \textbf{134} (2017)  135--141

\bibitem{leaf-disease2}
Ferentinos, K.P.:
\newblock Deep learning models for plant disease detection and diagnosis.
\newblock Comput. Electron. Agric. \textbf{145} (2018)  311--318

\bibitem{leaf-disease3}
Pallagani, V., Khandelwal, V., Chandra, B., Udutalapally, V., Das, D., Mohanty,
  S.P.:
\newblock dcrop: A deep-learning based framework for accurate prediction of
  diseases of crops in smart agriculture.
\newblock 2019 IEEE International Symposium on Smart Electronic Systems (iSES)
  (Formerly iNiS) (2019)  29--33

\bibitem{leaf-disease4}
Mohanty, S.P., Hughes, D.P., Salath{\'e}, M.:
\newblock Using deep learning for image-based plant disease detection.
\newblock In: Front. Plant Sci. (2016)

\bibitem{leaf-disease5}
Francis, M., Deisy, C.:
\newblock Disease detection and classification in agricultural plants using
  convolutional neural networks — a visual understanding.
\newblock 2019 6th International Conference on Signal Processing and Integrated
  Networks (SPIN) (2019)  1063--1068

\bibitem{rice1}
Li, D., Wang, R., Xie, C., Liu, L., Zhang, J., Li, R., Wang, F., Zhou, M., Liu,
  W.:
\newblock A recognition method for rice plant diseases and pests video
  detection based on deep convolutional neural network.
\newblock Sensors \textbf{20 3} (2020)

\bibitem{rice2}
jie Liang, W., Zhang, H., feng Zhang, G., xin Cao, H.:
\newblock Rice blast disease recognition using a deep convolutional neural
  network.
\newblock In: Scientific Reports. (2019)

\bibitem{rice3}
Zhou, G., Zhang, W., Chen, A., He, M., Ma, X.:
\newblock Rapid detection of rice disease based on fcm-km and faster r-cnn
  fusion.
\newblock IEEE Access \textbf{7} (2019)  143190--143206

\bibitem{tomato}
Maeda-Gutiérrez, V., Galván~Tejada, C., Zanella~Calzada, L., Celaya~Padilla,
  J., Galván~Tejada, J., Gamboa-Rosales, H., Luna-Garcia, H.,
  Magallanes-Quintanar, R., Carlos, G.M., Olvera-Olvera, C.:
\newblock Comparison of convolutional neural network architectures for
  classification of tomato plant diseases.
\newblock Applied Sciences \textbf{10} (02 2020)  1245

\bibitem{tomato2}
Fuentes, A., Yoon, S., Lee, J., Park, D.S.:
\newblock High-performance deep neural network-based tomato plant diseases and
  pests diagnosis system with refinement filter bank.
\newblock In: Front. Plant Sci. (2018)

\bibitem{tomato3}
Gutierrez, A., Ansuategi, A., Susperregi, L., Tubío, C., Rankić, I., Lenža,
  L.:
\newblock A benchmarking of learning strategies for pest detection and
  identification on tomato plants for autonomous scouting robots using internal
  databases.
\newblock Journal of Sensors \textbf{2019} (05 2019)  1--15

\bibitem{banana}
Michael Gomez~Selvaraj, Alejandro~Vergara, H.R.N.S.S.E.W.O.G.B.:
\newblock Ai-powered banana diseases and pest detection.
\newblock Plant Methods \textbf{2019} (08 2019)

\bibitem{grape}
Aravind, K.R., Raja, P., Aniirudh, R., Mukesh, K.V., Ashiwin, R., Vikas, G.:
\newblock Grape crop disease classification using transfer learning approach.
\newblock (2018)

\bibitem{sugarcane}
Shadab, M., Dwivedi, M., S~N, O., Javed, T., Bakey, A., Raqib, M.,
  Chakravarthy, A.:
\newblock Disease recognition in sugarcane crop using deep learning (09 2019)

\bibitem{eggplant}
Rangarajan, A.K., Purushothaman, R.:
\newblock Disease classification in eggplant using pre-trained vgg16 and msvm.
\newblock Scientific Reports \textbf{10} (2020)

\bibitem{cucumber}
Zhang, S., Wu, X., You, Z.H., Zhang, L.:
\newblock Leaf image based cucumber disease recognition using sparse
  representation classification.
\newblock Comput. Electron. Agric. \textbf{134} (2017)  135--141

\bibitem{soybean}
Kaur, S., Pandey, S., Goel, S.:
\newblock Semi-automatic leaf disease detection and classification system for
  soybean culture.
\newblock IET Image Processing \textbf{12} (2018)  1038--1048

\bibitem{olive}
Cruz, A.C., Luvisi, A., Bellis, L.D., Ampatzidis, Y.:
\newblock X-fido: An effective application for detecting olive quick decline
  syndrome with deep learning and data fusion.
\newblock In: Front. Plant Sci. (2017)

\bibitem{tea}
Chen, J., Liu, Q., Gao, L.:
\newblock Visual tea leaf disease recognition using a convolutional neural
  network model.
\newblock Symmetry \textbf{11} (2019)  343

\bibitem{coffee}
Esgario, J., Krohling, R., Ventura, J.:
\newblock Deep learning for classification and severity estimation of coffee
  leaf biotic stress (07 2019)

\bibitem{disease-detection-limits}
Arsenovic, M., Karanovic, M., Sladojevic, S., Anderla, A., Stefanovic, D.:
\newblock Solving current limitations of deep learning based approaches for
  plant disease detection.
\newblock Symmetry \textbf{11} (2019)  939

\bibitem{uav-survey1}
Shakhatreh, H., Sawalmeh, A.H., Al-Fuqaha, A., Dou, Z., Almaita, E., Khalil,
  I.M., Othman, N.S., Khreishah, A., Guizani, M.:
\newblock Unmanned aerial vehicles (uavs): A survey on civil applications and
  key research challenges.
\newblock IEEE Access \textbf{7} (2019)  48572--48634

\bibitem{uav-survey2}
Xiang, T., Xia, G.S., Zhang, L.:
\newblock Mini-unmanned aerial vehicle-based remote sensing: Techniques,
  applications, and prospects.
\newblock IEEE Geoscience and Remote Sensing Magazine \textbf{7} (2019)  29--63

\bibitem{uav-cropmanagement}
Huang, Y., Thomson, S.J., Hoffmann, W.C., Lan, Y., Fritz, B.K.:
\newblock Development and prospect of unmanned aerial vehicle technologies for
  agricultural production management.
\newblock (2013)

\bibitem{uav-cropmonitoring}
Dastgheibifard, S., Asnafi, M.:
\newblock A review on potential applications of unmanned aerial vehicle for
  construction industry.
\newblock (07 2018)

\bibitem{uav-weed-detect}
Kazmi, W., Bisgaard, M., Garcia-Ruiz, F.J., Hansen, K.D., la~Cour-Harbo, A.:
\newblock Adaptive surveying and early treatment of crops with a team of
  autonomous vehicles.
\newblock In: ECMR. (2011)

\bibitem{uav-irrigation}
V.~Gonzalez-Dugo, P. Zarco-Tejada, E.N.P.N.J.A.D.I., Fereres, E.:
\newblock Using high resolution uav thermal imagery to assess the variability
  in the water status of five fruit tree species within a commercial orchard.
\newblock Precision Agriculture \textbf{14}(6) (12 2013)  660--678

\bibitem{uav-agrovision}
Chiu, M.T., Xu, X., Wei, Y., Huang, Z., Schwing, A.G., Brunner, R.,
  Khachatrian, H., Karapetyan, H., Dozier, I., Rose, G., Wilson, D., Tudor,
  A.P., Hovakimyan, N., Huang, T.S., Shi, H.:
\newblock Agriculture-vision: A large aerial image database for agricultural
  pattern analysis.
\newblock ArXiv \textbf{abs/2001.01306} (2020)

\bibitem{uav-cattle}
Barbedo, J., Koenigkan, L., Santos, T., Santos, P.:
\newblock A study on the detection of cattle in uav images using deep learning.
\newblock Sensors \textbf{19} (12 2019)  5436

\bibitem{uav-disease}
Garcia-Ruiz, F., Sankaran, S., Maja, J.M., Lee, W.S., Rasmussen, J., Ehsani,
  R.:
\newblock Comparison of two aerial imaging platforms for identification of
  huanglongbing-infected citrus trees.
\newblock Computers and Electronics in Agriculture \textbf{91} (02 2013)
  106–115

\bibitem{uav-disease2}
Kerkech, M., Hafiane, A., Canals, R.:
\newblock Vine disease detection in uav multispectral images with deep learning
  segmentation approach.
\newblock (2019)

\bibitem{uav-insect}
Stumph, B., Virto, M.H., Medeiros, H., Tabb, A., Wolford, S., Rice, K., Leskey,
  T.C.:
\newblock Detecting invasive insects with unmanned aerial vehicles.
\newblock 2019 International Conference on Robotics and Automation (ICRA)
  (2019)  648--654

\bibitem{uav-datacollect}
Mathur, P., Nielsen, R.H., Prasad, N.R., Prasad, R.:
\newblock Data collection using miniature aerial vehicles in wireless sensor
  networks.
\newblock IET Wireless Sensor Systems \textbf{6} (2016)  17--25

\bibitem{uav-agromanagement}
Primicerio, J., Di~Gennaro, S., Fiorillo, E., Genesio, L., Lugato, E., Matese,
  A., Vaccari, F.:
\newblock A flexible unmanned aerial vehicle for precision agriculture.
\newblock Precision Agriculture (08 2012)

\bibitem{uav-irrigation-soil}
F.~M.~Rhoads, C.D.Y.:
\newblock Irrigation scheduling for corn—why and how.
\newblock National Corn Handbook (2000)

\bibitem{uav-soilpaper}
Hassan-Esfahani, L., Torres-Rua, A., Jensen, A., McKee, M.:
\newblock Assessment of surface soil moisture using high-resolution
  multi-spectral imagery and artificial neural networks.
\newblock Remote Sensing \textbf{7} (03 2015)  2627--2646

\bibitem{uav-soil-fungus}
Calderón~Madrid, R., Navas~Cortés, J., Lucena, C., Zarco-Tejada, P.:
\newblock High-resolution airborne hyperspectral and thermal imagery for early
  detection of verticillium wilt of olive using fluorescence, temperature and
  narrow-band spectral indices.
\newblock Remote Sensing of Environment \textbf{139} (09 2013)  231--245

\bibitem{uav-soiltexture1}
Wang, D.C., Zhang, G.L., Pan, X., Zhao, Y.G., Zhao, M.S., Wang, G.F.:
\newblock Mapping soil texture of a plain area using fuzzy-c-means clustering
  method based on land surface diurnal temperature difference.
\newblock Pedosphere \textbf{22} (06 2012)  394–403

\bibitem{uav-soiltexture2}
Wang, D.C., Zhang, G.L., Zhao, M.S., Pan, X., Zhao, Y.G., Li, D.C., Macmillan,
  B.:
\newblock Retrieval and mapping of soil texture based on land surface diurnal
  temperature range data from modis.
\newblock PLOS ONE \textbf{10} (06 2015)  e0129977

\bibitem{uav-soilresidue}
Sullivan, D., Shaw, J., Mask, P., Rickman, D., Guertal, E., Luvall, J.,
  Wersinger, J.:
\newblock Evaluation of multispectral data for rapid assessment of wheat straw
  residue cover.
\newblock Soil Science Society of America Journal \textbf{68} (11 2004)

\bibitem{uav-crop-maturity}
Jensen, T., Apan, A., Zeller, L.:
\newblock Crop maturity mapping using a low-cost low-altitude remote sensing
  system.
\newblock (2009)

\bibitem{uav-crop-yield1}
Swain, K., Thomson, S., Jayasuriya, H.:
\newblock Adoption of an unmanned helicopter for low-altitude remote sensing to
  estimate yield and total biomass of a rice crop.
\newblock Transactions of the ASABE \textbf{53} (01 2010)

\bibitem{uav-crop-yield2}
Geipel, J., Link, J., Claupein, W.:
\newblock Combined spectral and spatial modeling of corn yield based on aerial
  images and crop surface models acquired with an unmanned aircraft system.
\newblock Remote Sensing \textbf{6} (11 2014)  10335--10355

\bibitem{uav-ground}
Pretto, A., Aravecchia, S., Burgard, W., Chebrolu, N., Dornhege, C., Falck, T.,
  Fleckenstein, F.V., Fontenla, A., Imperoli, M., Khanna, R., Liebisch, F.,
  Lottes, P., Milioto, A., Nardi, D., Nardi, S., Pfeifer, J., Popovic, M.,
  Potena, C., Pradalier, C., Rothacker-Feder, E., Sa, I., Schaefer, A.,
  Siegwart, R., Stachniss, C., Walter, A., Winterhalter, W., Wu, X.L., Nieto,
  J.:
\newblock Building an aerial-ground robotics system for precision farming.
\newblock ArXiv \textbf{abs/1911.03098} (2019)

\bibitem{uav-conclusion}
Primicerio, J., Di~Gennaro, S., Fiorillo, E., Genesio, L., Lugato, E., Matese,
  A., Vaccari, F.:
\newblock A flexible unmanned aerial vehicle for precision agriculture.
\newblock Precision Agriculture (08 2012)

\bibitem{satellites-gamaya}
Press:
\newblock How satellites are making agriculture more efficient.
\newblock Gamaya Blog Post (2017)

\bibitem{satellites-india}
Press:
\newblock Satellites designed for benefit of farmers.
\newblock Government of India, Department of Space (2016)

\bibitem{esa-earth-online2}
Unknown:
\newblock Farmers benefit from satellite coverage.
\newblock ESA Earth Online

\bibitem{satellite-crop-type}
Ru{\ss}wurm, M., Lef{\`e}vre, S., K{\"o}rner, M.:
\newblock Breizhcrops: A satellite time series dataset for crop type
  identification.
\newblock ArXiv \textbf{abs/1905.11893} (2019)

\bibitem{satellite-soil3}
Ghazali, M., Wikantika, K., Harto, A., Kondoh, A.:
\newblock Generating soil salinity, soil moisture, soil ph from satellite
  imagery and its analysis.
\newblock Information Processing in Agriculture (08 2019)

\bibitem{satellite-soil1}
Sheffield, K., Morse-McNabb, E.:
\newblock Using satellite imagery to asses trends in soil and crop productivity
  across landscapes.
\newblock IOP Conference Series: Earth and Environmental Science \textbf{25}
  (07 2015)  012013

\bibitem{satellite-soil2}
Kumar, N., Anouncia, S., Madhavan, P.:
\newblock Application of satellite remote sensing to find soil fertilization by
  using soil colour.
\newblock International Journal of Online Engineering \textbf{9} (05 2013)

\bibitem{geocento}
Geocento:
\newblock Online provider of satellite and drone imagery

\bibitem{satellite-imagery-sources}
Unknown:
\newblock 7 top free satellite imagery sources in 2019.
\newblock Earth Observatory System (2019)

\bibitem{esa-earth-online}
Unknown:
\newblock Agriculture overview.
\newblock ESA Earth Online

\bibitem{data-aug-arigan}
Giuffrida, M.V., Scharr, H., Tsaftaris, S.A.:
\newblock Arigan: Synthetic arabidopsis plants using generative adversarial
  network.
\newblock 2017 IEEE International Conference on Computer Vision Workshops
  (ICCVW) (2017)  2064--2071

\bibitem{gans}
Goodfellow, I.J., Pouget-Abadie, J., Mirza, M., Xu, B., Warde-Farley, D.,
  Ozair, S., Courville, A.C., Bengio, Y.:
\newblock Generative adversarial networks.
\newblock ArXiv \textbf{abs/1406.2661} (2014)

\bibitem{data-aug-rosette1}
Zhu, Y., Aoun, M., Krijn, M., Vanschoren, J.:
\newblock Data augmentation using conditional generative adversarial networks
  for leaf counting in arabidopsis plants.
\newblock In: BMVC. (2018)

\bibitem{data-aug-rosette2}
Kuznichov, D., Zvirin, A., Honen, Y., Kimmel, R.:
\newblock Data augmentation for leaf segmentation and counting tasks in rosette
  plants.
\newblock In: CVPR Workshops. (2019)

\bibitem{data-aug-unsup-gan}
Nazki, H., Yoon, S., Fuentes, A., Park, D.S.:
\newblock Unsupervised image translation using adversarial networks for
  improved plant disease recognition.
\newblock Comput. Electron. Agric. \textbf{168} (2019)

\bibitem{leafgan}
Cap, Q.H., Uga, H., Kagiwada, S., Iyatomi, H.:
\newblock Leafgan: An effective data augmentation method for practical plant
  disease diagnosis.
\newblock (2020)

\bibitem{data-enhance-disease}
Sun, R., Zhang, M., Yang, K.:
\newblock Data enhancement for plant disease classification using generated
  lesions.
\newblock (2020)

\bibitem{data-aug-miltimodal}
Sapoukhina, N., Samiei, S., Rasti, P., Rousseau, D.:
\newblock Data augmentation from rgb to chlorophyll fluorescence imaging
  application to leaf segmentation of arabidopsis thaliana from top view
  images.
\newblock In: CVPR Workshops. (2019)

\bibitem{weak-sup2}
Bellocchio, E., Ciarfuglia, T., Costante, G., Valigi, P.:
\newblock Weakly supervised fruit counting for yield estimation using spatial
  consistency.
\newblock IEEE Robotics and Automation Letters \textbf{PP} (03 2019)  1--1

\bibitem{weak-sup3}
Marino, S., Beauseroy, P., Smolarz, A.:
\newblock Weakly-supervised learning approach for potato defects segmentation.
\newblock Engineering Applications of Artificial Intelligence \textbf{85} (07
  2019)  337--346

\bibitem{Desai2019AnAS}
Desai, S.V., Chandra, A.L., Guo, W., Ninomiya, S., Balasubramanian, V.N.:
\newblock An adaptive supervision framework for active learning in object
  detection.
\newblock British Machine Vision Conference (2019)

\bibitem{transfer-learning-2}
Mehdipour-Ghazi, M., Yanikoglu, B.A., Aptoula, E.:
\newblock Plant identification using deep neural networks via optimization of
  transfer learning parameters.
\newblock Neurocomputing \textbf{235} (2017)  228--235

\bibitem{imagenet}
Deng, J., Dong, W., Socher, R., Li, L.J., Li, K., Li, F.F.:
\newblock Imagenet: a large-scale hierarchical image database.
\newblock (06 2009)  248--255

\bibitem{lifeclef}
Joly, A., Goëau, H., Spampinato, C., Bonnet, P., Vellinga, W.P., Planqué, R.,
  Rauber, A., Palazzo, S., Fisher, B., Müller, H.:
\newblock Lifeclef 2015: Multimedia life species identification challenges.
\newblock (09 2015)

\bibitem{transfer-learning-1}
Xu, W., Yu, G., Zare, A., Zurweller, B., Rowland, D., Reyes-Cabrera, J.,
  Fritschi, F.B., Matamala, R., Juenger, T.E.:
\newblock Overcoming small minirhizotron datasets using transfer learning.
\newblock ArXiv \textbf{abs/1903.09344} (2019)

\bibitem{tl-leaf-disease}
Malpe, S.:
\newblock Automated leaf disease detection and treatment recommendation using
  transfer learning.
\newblock (2019)

\bibitem{domain-adapt}
Bellocchio, E., Costante, G., Cascianelli, S., Fravolini, m., Valigi, P.:
\newblock Combining domain adaptation and spatial consistency for unseen fruits
  counting: A quasi-unsupervised approach.
\newblock IEEE Robotics and Automation Letters \textbf{5} (01 2020)  1--1

\bibitem{ch2019active}
Chandra, A.L., Desai, S.V., Balasubramanian, V.N., Ninomiya, S., Guo, W.:
\newblock Active learning with point supervision for cost-effective panicle
  detection in cereal crops.
\newblock BMC Plant Methods (2020)

\bibitem{Settles10activelearning}
Settles, B.:
\newblock Active learning literature survey.
\newblock Technical report, University of Wisconsin–Madison (2010)

\bibitem{Gal2017DeepBA}
Gal, Y., Islam, R., Ghahramani, Z.:
\newblock Deep bayesian active learning with image data.
\newblock In: ICML. (2017)

\bibitem{Sener2018ActiveLF}
Sener, O., Savarese, S.:
\newblock Active learning for convolutional neural networks: A core-set
  approach.
\newblock In: ICLR 2018. (2018)

\bibitem{Wang_gupta}
Wang, K., Zhang, D., Li, Y., Zhang, R., Lin, L.:
\newblock Cost-effective active learning for deep image classification.
\newblock IEEE Trans. Cir. and Sys. for Video Technol. \textbf{27}(12)
  (December 2017)  2591--2600

\bibitem{al-deep-object-baseline}
Brust, C., K{\"{a}}ding, C., Denzler, J.:
\newblock Active learning for deep object detection.
\newblock CoRR \textbf{abs/1809.09875} (2018)

\bibitem{VinayNamboodiri2018}
Roy, S., Unmesh, A., Namboodiri, V.P.:
\newblock Deep active learning for object detection.
\newblock In: BMVC. (2018)

\bibitem{Vijayanarasimhan2014}
Vijayanarasimhan, S., Grauman, K.:
\newblock Large-scale live active learning: Training object detectors with
  crawled data and crowds.
\newblock International Journal of Computer Vision \textbf{108}(1) (May 2014)
  97--114

\end{thebibliography}
\end{document}